\documentclass{article}

\usepackage{arxiv}
\usepackage{amsmath} 
\usepackage[utf8]{inputenc} 
\usepackage[T1]{fontenc}    
\usepackage{url}            
\usepackage{booktabs}       
\usepackage{amsfonts}       
\usepackage{nicefrac}       
\usepackage{microtype}      
\usepackage{lipsum}
\usepackage{graphicx}
\graphicspath{ {./images/} }
\usepackage{tikz}
\usetikzlibrary{positioning, arrows.meta, shapes, calc, decorations.pathreplacing}

\usepackage{enumitem}

\usepackage{amsmath}
\usepackage{algorithm}
\usepackage{algorithmic}

\usepackage{graphicx}
\usepackage{caption}
\usepackage{subcaption}
\usepackage{appendix}
\usepackage{float}

\usepackage{multirow}
\usepackage{amsmath}
\usepackage{amssymb}

\title{AI-Driven Multi-Stage Computer Vision System for Defect Detection in Laser-Engraved Industrial Nameplates}

\author{
  Adhish Anitha Vilasan \\
  Master's Student (Lead Author)\\
  Technische Hochschule Rosenheim\\
  Rosenheim, Germany \\
  \texttt{adhish.anithavilasan@stud.th-rosenheim.de} \\
  \And
  Stephan Jäger \\
  Data Scientist\\
  Knorr-Bremse Systeme für Nutzfahrzeuge GmbH\\
  Aldersbach, Germany \\
  \texttt{Stephan.Jaeger@knorr-bremse.com} \\
  \And
  Noah Klarmann \\
  Faculty of Management and Engineering\\
  Technische Hochschule Rosenheim\\
  Rosenheim, Germany \\
  \texttt{noah.klarmann@th-rosenheim.de} \\
}

\begin{document}
\newcommand{\none}{}
\maketitle

\begin{abstract}
Automated defect detection in industrial manufacturing is essential for maintaining product quality and minimizing production errors. In air disc brake manufacturing, ensuring the precision of laser-engraved nameplates is crucial for accurate product identification and quality control. Engraving errors, such as misprints or missing characters, can compromise both aesthetics and functionality, leading to material waste and production delays. This paper presents a proof of concept for an AI-driven computer vision system that inspects and verifies laser-engraved nameplates, detecting defects in logos and alphanumeric strings. The system integrates object detection using YOLOv7, optical character recognition (OCR) with Tesseract, and anomaly detection through a residual variational autoencoder (ResVAE) along with other computer vision methods to enable comprehensive inspections at multiple stages. Experimental results demonstrate the system’s effectiveness, achieving 91.33\% accuracy and 100\% recall, ensuring that defective nameplates are consistently detected and addressed. This solution highlights the potential of AI-driven visual inspection to enhance quality control, reduce manual inspection efforts, and improve overall manufacturing efficiency.
\end{abstract}

\section{Introduction}
In the manufacturing of air disc brake systems, laser engraving of nameplates is a critical process that embeds essential details such as customer names, logos, production data, Data Matrix Code (DMC), and other alphanumeric characters. Errors during engraving are often caused by the accumulation of metallic dust and debris generated during the process, which settles on the laser housing. This buildup causes the laser to hit the obstacle instead of the nameplate, leading to misprints or incomplete markings. Such defects not only compromise the visual and professional quality of the nameplates but also create significant challenges in traceability and accurate product identification, which are essential for addressing safety concerns, ensuring production accountability, and managing batch-related claims. Early detection of these defects, before the nameplates are affixed to the disc brakes, is crucial for maintaining production efficiency, avoiding disruptions in the manufacturing process, and ensuring that only defect-free nameplates reach customers.

\begin{figure}[h!]
    \centering
    \begin{subfigure}[t]{0.45\textwidth}
        \centering
        \includegraphics[width=\textwidth]{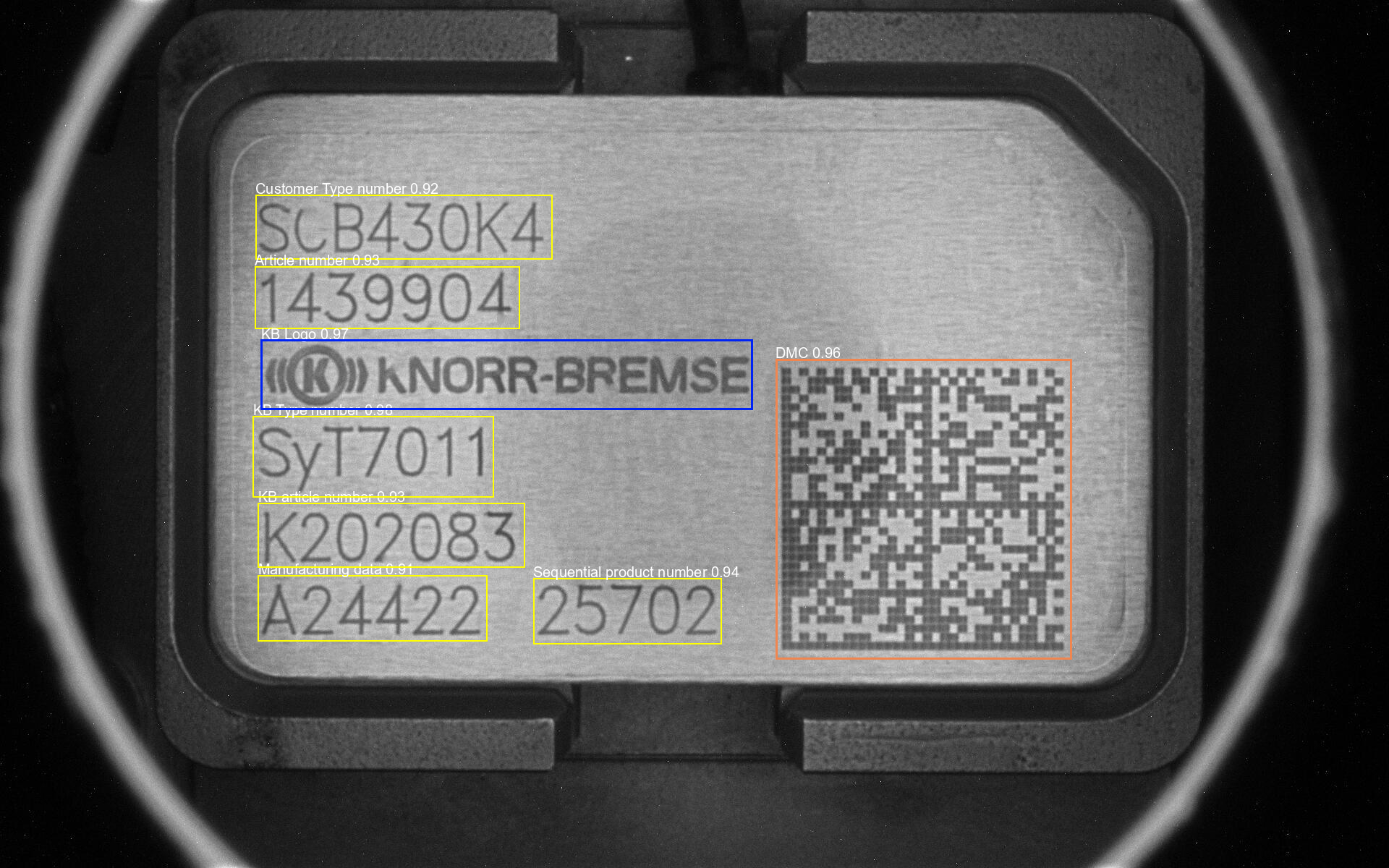}
        \caption{}
        \label{fig:image1}
    \end{subfigure}
    \hspace{0.5em} 
    \begin{subfigure}[t]{0.45\textwidth}
        \centering
        \includegraphics[width=\textwidth]{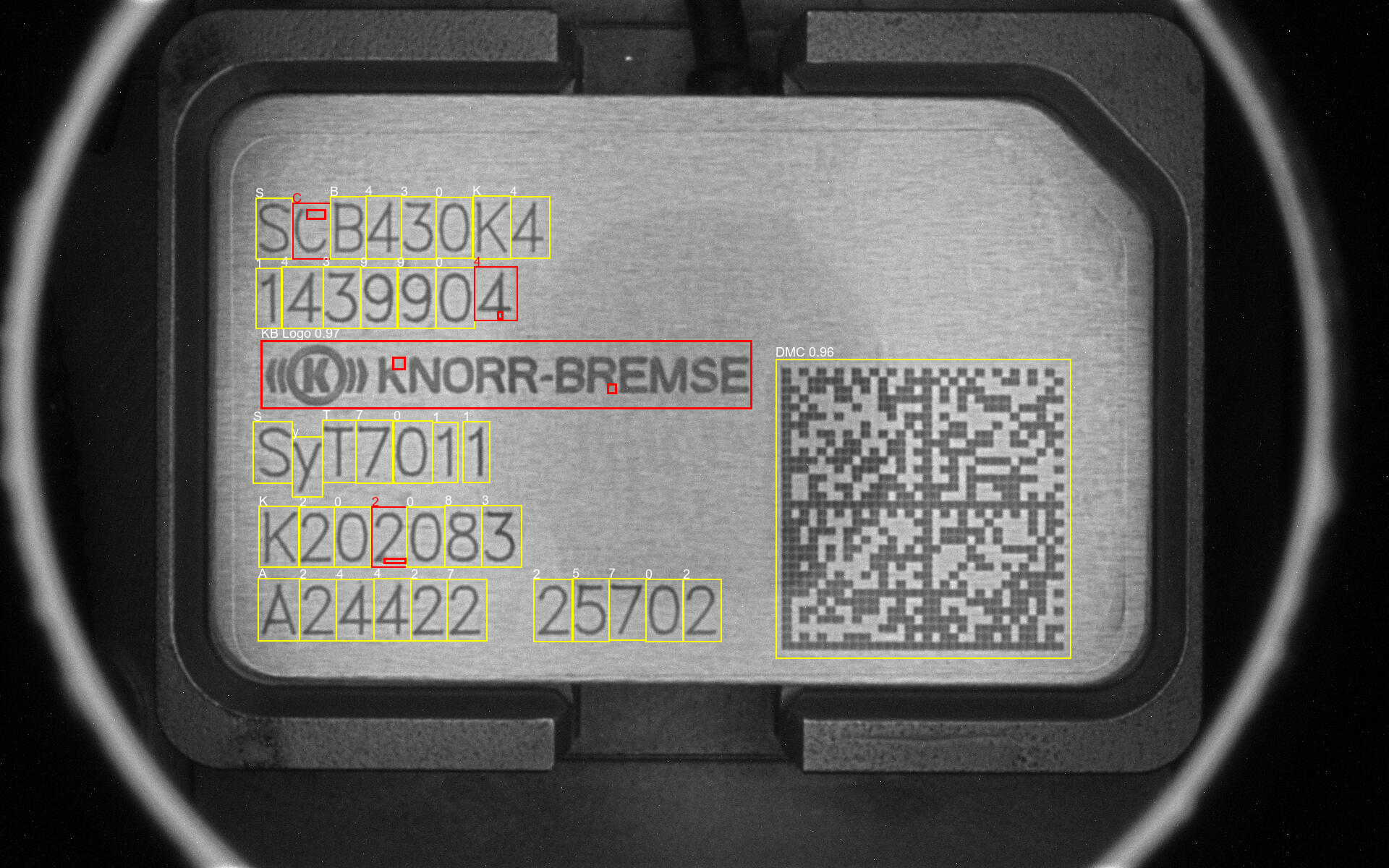}
        \caption{}
        \label{fig:image2}
    \end{subfigure}
    \caption{(a) Annotated defective laser-engraved result from YOLOv7 \cite{wang2023yolov7}, with a blue bounding box for the logo and yellow bounding boxes for strings, along with confidence values, (b) Final pipeline output with defects highlighted in red boxes.}
    \label{fig:combined_images}
\end{figure}

Due to varying customer requirements, nameplate layouts differ across air disc brake systems, making laser engraving necessary for each design. Each nameplate is unique, differing in the content and length of the strings engraved on it. An example of this heterogeneity is shown in Figure \ref{fig:heterogeneity_data}. As a result, thorough inspection of these nameplates is essential to ensure their quality. An example of such defects is presented in Figure \ref{fig:combined_images}(a), which highlights errors in engraved logos and strings. 

For the purpose of developing the inspection pipeline, the laser-engraved content is categorized into two sections: logos and strings. As shown in Figure \ref{fig:combined_images}(a), the "KNORR-BREMSE" logo, enclosed within a bounding box drawn in blue, is classified under logos. Additionally, customer-specific logos and names are also engraved but are not shown in this paper due to confidentiality. All other textual elements on the nameplates are classified as strings. It is important to note that checking the quality of the Data Matrix Code (DMC), which is enclosed in a bounding box drawn in orange, is not within the scope of this paper and was not considered during the development of the inspection pipeline, as ready-to-use solutions already exist, making further research and development unnecessary. 

This paper focuses on designing an AI-based computer vision system tailored to inspect nameplates during air disc brake manufacturing. The system integrates object detection, optical character recognition (OCR), anomaly detection, and traditional computer vision techniques to thoroughly analyze both text and image content on nameplates.

The primary objective of the proposed system is to ensure the precision and completeness of laser engravings. This involves verifying that logos and characters are accurately replicated without defects. Additionally, the system performs cross-referencing of extracted nameplate information against data provided by the Manufacturing Execution System (MES). This verification step is critical to ensure the consistency and correctness of the engraved information.

Furthermore, the proof of concept (PoC) aims to generalize across various nameplate layouts, effectively detecting defects regardless of design variations. The developed pipeline not only identifies print defects but also localizes them on the nameplate. This simplifies defect isolation and facilitates corrective actions.

\section{Related Works}

\noindent In recent years, automated visual inspection in industrial manufacturing has gained significant traction as a strategy to enhance product quality, reduce downtime, and minimize reliance on manual checks \cite{kim2018smart, liao2019state}. Early inspection solutions primarily relied on traditional computer vision techniques such as thresholding, morphological operations, and template matching. While these rule-based methods can be effective under strictly controlled conditions, they often prove insufficient in settings with high variability in lighting, part orientation, or background noise \cite{xie2008review, aslam2020comparative, kumar2008computer}. Nonetheless, they remain useful in certain stages of modern inspection pipelines. For instance, Ma \textit{et al.} \cite{8278352} proposed a two-stage template matching approach that first identifies regions of interest in personalized prints, followed by thresholding and morphological filters to isolate potential defects. Their method demonstrated the viability of template matching for intricate, irregular surfaces but also highlighted the need for robust pre-processing to accommodate variations in background and foreground regions.  Raihan and Ce (2017) \cite{ce2017pcb} employs traditional computer vision techniques, including image subtraction, Gaussian blurring, thresholding, and blob detection in OpenCV, to detect PCB defects by comparing reference and inspected images, achieving rapid processing times. However, such approaches still require manual adjustments for dynamic layouts, limiting their applicability to variable industrial tasks. To address these challenges, modern systems increasingly integrate traditional methods with deep learning, balancing computational efficiency with adaptability to complex industrial environments.

\noindent \textbf{Deep Learning in Industrial Defect Detection.} 
Deep learning has transformed defect detection by automating feature extraction from large datasets \cite{yang2020using}. For example, Li \textit{et al.} \cite{9873703} introduced a ResNet-based approach for defect classification in printed products, achieving an accuracy, precision, and recall exceeding 96\% across five defect types. Their research highlights the effectiveness of deep learning integration in machine vision systems for handling intricate printing defects influenced by variables such as equipment, environmental conditions, and printing technology. Object detection architectures, such as Faster R-CNN \cite{7485869} and Detectron2 \cite{Detectron2}, further advanced industrial applications by identifying objects or regions of interest in a single pass. The YOLO family of detectors \cite{wang2023comprehensive} further enhanced both speed and accuracy, making them particularly suitable for industrial environments with stringent cycle time constraints. By utilizing a single neural network to analyze an entire image, YOLO divides it into multiple regions and predicts bounding boxes and class probabilities for each segment \cite{wang2023yolov7}. Its ability to maintain a balance between detection precision and real-time processing makes YOLO a preferred choice in manufacturing applications \cite{sharma2024optimizing}.

\noindent \textbf{Optical Character Recognition (OCR) Method.} 
Verifying the accuracy of engraved or printed text is essential for product identification and traceability in manufacturing. Over time, Optical Character Recognition (OCR) techniques have progressed from traditional hand-engineered feature extraction methods to deep learning-based approaches \cite{raj2022comprehensive}. Among these, Tesseract OCR \cite{smith2007overview}, an open-source engine developed by Google, remains widely utilized due to its support for multiple languages and dependable performance in structured text environments \cite{anwar2022text}. It processes text at the character level by combining pattern recognition with feature extraction and machine learning techniques, such as Long Short-Term Memory (LSTM) networks, to accurately identify individual characters \cite{smith2007overview}. Although Tesseract can struggle with noise, varying backgrounds, and text distortions, studies have demonstrated that targeted pre-processing can significantly improve its accuracy. For instance, Brisinello \textit{et al.} \cite{brisinello2017improving} improved Tesseract 4.0’s accuracy from 70.2\% to 92.9\% by applying pre-processing techniques such as image resizing, sharpening, blurring, and foreground-background separation using k-means clustering.

Beyond Tesseract, several advanced OCR frameworks present alternative approaches. Li \textit{et al.} \cite{li2023trocr} introduced TrOCR, a Transformer-based model that eliminates CNNs and RNNs in favor of pre-trained vision and language models (e.g., ViT/BEiT and RoBERTa/MiniLM), achieving state-of-the-art performance in printed, handwritten, and scene text recognition. However, TrOCR does not support character-level detection. Character-centric frameworks such as WordSup \cite{hu2017wordsup} and CRAFT \cite{baek2019character} address this limitation. WordSup utilizes word-level annotations to train character detectors with a VGG16-FPN network, whereas CRAFT employs a VGG16-U-Net hybrid to predict character regions and affinity scores. These methods excel at detecting curved and multi-oriented text but focus solely on detection rather than recognition and require significant computational resources.

\noindent \textbf{Anomaly Detection Method}
In industrial settings, unsupervised methods are particularly valuable when defect samples are rare or poorly defined \cite{tao2022deep}. Reconstruction-based anomaly detection relies on generative models like autoencoders (AEs) or generative adversarial networks (GANs) \cite{goodfellow2014generative}. These models learn to reconstruct normal images from defect-free training data, enabling defect detection by comparing original and reconstructed images. GANs typically produce sharper reconstructions but are harder to train, while AEs offer simpler training at the cost of blurrier outputs. For example, Chow \textit{et al.} \cite{chow2020anomaly} proposed an unsupervised convolutional autoencoder to detect defects in concrete structures, using high reconstruction errors to localize anomalies. Similarly, Shi \textit{et al.} \cite{shi2023lightweight} introduced a lightweight reconstruction network (LRN-L) for surface defect detection, leveraging residual analysis and adaptive thresholding to reduce false positives. However, traditional autoencoders often overfit to training data and struggle to generalize to unseen anomalies \cite{kingma2013auto, chen2023auto}.

Variational Autoencoders (VAEs) address these limitations by framing the encoding process as a Bayesian inference problem \cite{kingma2013auto}. Instead of mapping inputs to fixed latent vectors, VAEs encode them into probability distributions, balancing reconstruction quality with latent-space regularization. However, VAEs often produce blurry reconstructions, particularly for complex or irregular patterns \cite{he2022survey}. Recent advancements have improved reconstruction fidelity by incorporating perceptual loss functions and residual connections. For instance, Hou \textit{et al.} \cite{xianxu2024feature} replaced pixel-wise loss with feature perceptual loss using pre-trained CNNs (e.g., VGG), aligning high-level semantic features for clearer reconstructions. Kumar \textit{et al.} \cite{10104785} enhanced VAEs with residual blocks, improving gradient flow and enabling the model to capture intricate patterns in video data. Their Residual VAE (RVAE) outperformed standard VAEs and convolutional autoencoders in detection accuracy and reconstruction quality.

Recent methods have extended VAE-based approaches to multi-class anomaly detection. You \textit{et al.} \cite{you2022unified} proposed UniAD, a unified framework that mitigates the "identical shortcut" issue through layer-wise query decoding, neighbor-masked attention, and feature jittering. Similarly, He \textit{et al.} \cite{he2024diffusion} introduced DiAD, a diffusion-based approach that integrates a Semantic-Guided (SG) network and Spatial-aware Feature Fusion (SFF) block to enhance reconstruction accuracy while preserving semantic integrity. These methods achieve state-of-the-art performance on benchmark datasets like MVTec-AD and VisA, demonstrating the potential of advanced anomaly detection techniques for industrial applications.

\noindent \textbf{Practical Advances in Industrial Defect Detection}  
Several researchers have introduced specialized systems tailored to defect detection and recognition across varied industrial settings. For instance, Li \textit{et al.} \cite{6999108} developed a machine vision–based approach for nameplate inspection, combining seed algorithms for pre-processing, BLOB analysis for boundary checks, and grid-based template matching with Euclidean distance metrics—ultimately achieving rapid processing times and high accuracy compared to manual methods. Peng \textit{et al.} \cite{peng2021defect} proposed the BBE framework, which applies morphological operations (e.g., erosion/dilation) and data augmentation to handle code characters on complex surfaces, employing EfficientNet and a two-part architecture (BUNet and BWNet) for multi-scale fusion. Similarly, Liu \textit{et al.} \cite{liu2023printing} introduced a scale-adaptive template matching method enhanced by CNN-based feature extraction and image alignment, achieving 93.62\% accuracy in printing defect detection by correcting positional deviations.

In another work, Elanangai \textit{et al.} \cite{elanangai2019automated} presented an automated system for stainless steel plate inspection, leveraging Multi-Scale LoG Weighting, SVM-RFE classification, and pattern correlation to reach 94.88\% accuracy—surpassing alternative methods like Random Forest and ANN. More recently, Xu \textit{et al.} \cite{haitao2023surface} proposed an enhanced YOLOv5 model for bearing ring surface defect detection, incorporating C2f modules, SPD layers, and CARAFE upsampling. Their modifications yielded a 97.3\% mAP and 100 FPS on an industrial dataset, outperforming standard YOLOv5 and YOLOv7.

\noindent While existing research has made significant strides in object detection, OCR, and anomaly detection, few studies address these components in a single, fully integrated pipeline. Moreover, none of the works cited in this section specifically target the complexities of laser-engraved nameplates that exhibit widely varying layouts and content. The method proposed here aims to fill this gap by integrating state-of-the-art detection model, OCR technique, and anomaly detection with a dedicated content validation step. This holistic design not only identifies print defects in both logos and text, but also cross-checks engraved information against manufacturing data. By accommodating diverse engraving patterns and enabling efficient defect isolation, the system strengthens product traceability and supports high manufacturing standards, offering a comprehensive solution tailored to the demands of air disc brake production.

\begin{figure}[h!]
  \centering
  \includegraphics[width=0.95\textwidth]{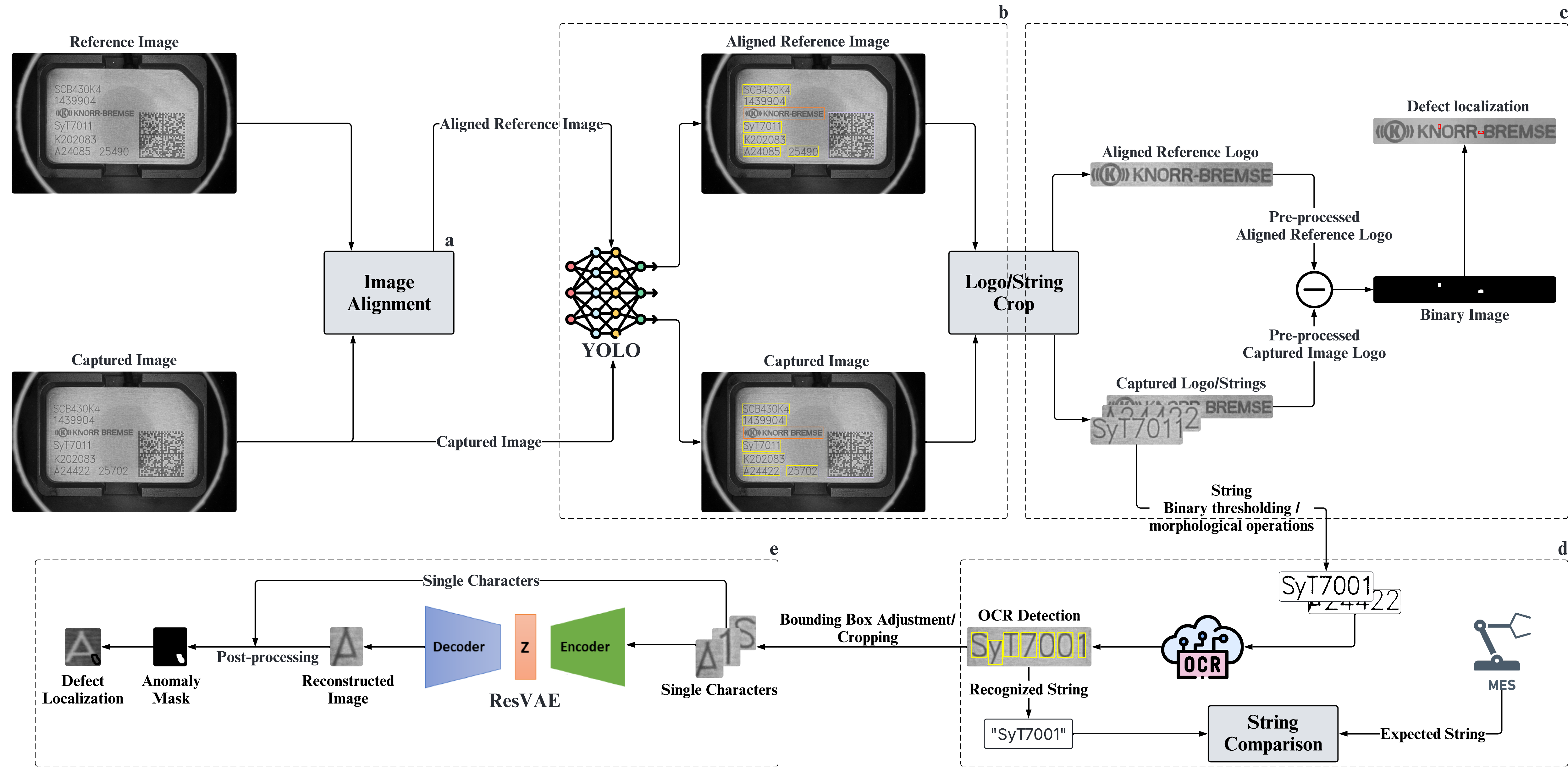}
  \caption{The complete pipeline for nameplate defect detection and string verification. 
    (a) Image alignment, (b) YOLO-based string and logo detection, (c) Logo defect detection through image differencing, (d) OCR for character recognition/detection, and (e) Anomaly detection using ResVAE.}
  \label{fig:pipeline}
\end{figure}

\section{Methodology}
\label{sec:methodology}

The proposed method employs multiple models to detect print defects and mismatches in the laser-engraved content on nameplates, including both strings and logos. By leveraging specialized models and techniques for distinct tasks, the system enhances defect detection and localizes defects at the character level, resulting in improved accuracy and performance.

\subsection{Overview of the Pipeline}

The developed pipeline integrates several models and computer vision techniques to inspect laser-engraved content for defects. A YOLOv7 object detection model is utilized to detect and classify each string and logo on the nameplate. For character detection and recognition, Tesseract OCR is applied. Defect detection for logos and characters follows two distinct approaches: traditional computer vision techniques are employed for logo defect detection, while a variational autoencoder (VAE) with residual blocks in the encoder and decoder is used for character-level anomaly detection.

Each component in the pipeline serves a specific role in verifying the accuracy of the laser-engraved content on the nameplate. The complete pipeline, along with the sequence of models and techniques used, is illustrated in Figure \ref{fig:pipeline}. The inferencing steps of the pipeline are detailed in Algorithm \ref{alg:nameplate_inspection}. 

The pipeline consists of multiple tasks, each with specific inputs, outputs, and dependencies. These dependencies ensure that the output of one task provides the necessary input for the subsequent task, creating a structured and efficient workflow. The tasks and their dependencies are outlined below:

\begin{enumerate}[label=\alph*.]
    \item \textbf{Image Alignment}  
    \begin{itemize}
        \item \textbf{Input:} Reference image and captured image.  
        \item \textbf{Output:} Aligned reference image.  
        \item \textbf{Dependency:} Image alignment is essential for downstream tasks, as it ensures the reference image and the captured image are spatially aligned. This step is particularly critical for accurate string and logo detection, which rely on precise alignment to minimize localization errors.
    \end{itemize}
    \item \textbf{YOLO String and Logo Detection}  
    \begin{itemize}
        \item \textbf{Input:} Aligned reference image and captured image.  
        \item \textbf{Output:} Cropped logo from the reference image and cropped strings and logos from the captured image.  
        \item \textbf{Dependency:} Requires aligned images from the image alignment step to ensure the bounding boxes of the detected logos and strings correspond accurately to their positions in the reference image.
    \end{itemize}
    \item \textbf{Logo Defect Detection}  
    \begin{itemize}
        \item \textbf{Input:} Cropped logo from the reference image and cropped logo from the captured image.  
        \item \textbf{Output:} Classification of the logo as acceptable or defective.  
        \item \textbf{Dependencies:} Relies on the output from the YOLO \cite{wang2023yolov7} detection model and image alignment to ensure that the comparison is performed between the corresponding logo regions.
    \end{itemize}
    \item \textbf{Tesseract OCR for Character Detection and Recognition}  
    \begin{itemize}
        \item \textbf{Input:} Cropped strings from the captured image.  
        \item \textbf{Output:} Individual characters recognized from the strings.  
        \item \textbf{Dependencies:} Requires cropped strings from YOLO detection and expected strings retrieved from the Manufacturing Execution System (MES) to compare recognized characters with expected values.
    \end{itemize}
    \item \textbf{Anomaly Detection Using ResVAE}  
    \begin{itemize}
        \item \textbf{Input:} Individual characters detected by Tesseract OCR \cite{smith2007overview}.  
        \item \textbf{Output:} Classification of characters as acceptable or defective.  
        \item \textbf{Dependencies:} Requires individual characters as input from Tesseract OCR to perform character-level anomaly detection.
    \end{itemize}
\end{enumerate}

Each task in the pipeline builds upon the output of the preceding tasks, highlighting the interdependencies between different components. A detailed explanation of each stage, along with experimental evaluations and results, is provided in the subsequent sections.

\subsection{Dataset Preparation}
\label{sec:dataset}

The development and evaluation of the pipeline relied on multiple datasets, each tailored to the specific requirements of different stages of the pipeline. All nameplates used for development had a consistent layout, with fixed positions for laser-engraved strings and logos, as well as a consistent string length. However, the content of the strings varied across nameplates. Examples of this variation are shown in Figure \ref{fig:heterogeneity_data}(a) and (b).

High-quality data captured from the production line were carefully selected to ensure robust training and accurate evaluation of the models. The selection criteria required that images be free from distortions such as poor lighting, low contrast, or excessive blur, which would make it difficult for humans to confidently detect the objects. Images that did not meet these criteria were excluded. All images used in this study were captured in grayscale using a monochrome camera on the production line, resulting in single-channel images. Due to the limited availability of defective images, all defective images used for training and testing were manually created using an image manipulation tool called GIMP \cite{cutts1997introduction}.

\begin{table}[h]
\centering
\caption{Train-Test Split}
\label{tab:pipeline}
\renewcommand{\arraystretch}{1.3} 
\begin{tabular}{l l c c c}
\hline
\multicolumn{2}{c}{\textbf{Stage}}                                  & \multicolumn{2}{c}{\textbf{Images}}            \\ 
\hline
\textbf{Name}                     & \textbf{Model/Method}               & \textbf{Train} & \textbf{Test} & \textbf{Size (pixels)} \\ 
\hline
String Detection                  & YOLOv7                                & 476               & 65          & 1920x1600             \\ 
Logo Defect Detection             & Image Differencing                    & -                 & 115         & 1920x1600             \\ 
Character Detection/Recognition   & Tesseract OCR (Pre-trained)           & -                 & 65          & 1920x1600             \\ 
Anomaly Detection on Characters   & Residual Variational Autoencoder      & 2957              & 500         & 64x64                 \\ 
Pipeline Testing/Evaluation       & Full Pipeline Testing                 & -                 & 150         & 1920x1600             \\ 
\hline
\end{tabular}
\end{table}

Although the dataset sizes in Table~\ref{tab:pipeline} may appear limited, they reflect practical constraints in a production environment. Defective nameplates are relatively rare, which makes it difficult to collect a large number of truly defective samples. To address this limitation, some defective examples were manually synthesized to broaden coverage of potential anomalies. Furthermore, training data for YOLO also required substantial annotation time, as each image had to be reviewed to confirm accurate identification of text and logo regions. Despite the smaller volume, each subset of data was carefully curated for quality, consistency, and coverage of the most common variation factors, including lighting changes and minor positional shifts, to maintain model robustness.

Table~\ref{tab:pipeline} summarizes the distribution of images across the pipeline stages. The complete 1920×1600-pixel nameplate image, as shown in Figure \ref{fig:image1}, was used for evaluation in most pipeline stages. However, for training and testing the Residual Variational Autoencoder, individual characters were extracted and resized to 64×64 pixels, as this is the input size required by the model. If the extracted characters were not square, they were resized to meet the required dimensions before training. Figure~\ref{fig:character_example} illustrates examples of all characters used for training the ResVAE, while Figure~\ref{fig:manu_character_example} presents a few synthetically generated defective characters.

To ensure robustness during testing, the datasets included images captured under varying lighting conditions, as illustrated in Figure \ref{fig:lighting_variation}. Dedicated training datasets were required for both the YOLO-based string and logo detection modules and the anomaly detection network. Annotated images were used to train and evaluate the YOLO model, ensuring accurate detection of strings and logos. In contrast, components such as Tesseract OCR utilized pre-trained models and did not require additional training. Similarly, the logo defect detection module, which is not based on machine learning, did not require any training dataset.

\subsection{Image Alignment and Logo Defect Detection}
\label{subsec: Image Alignment and Logo Defect Detection}
\subsubsection{Method and Process}

The primary objective of this method is to detect and locate defects in the logos of the nameplate. A traditional computer vision approach is employed, ensuring simplicity and efficiency in implementation.

The logo defect detection process is divided into two stages: image alignment and defect detection using conventional computer vision methods.

\paragraph{Image Alignment}
\label{subsec: Image alignment}

Image alignment plays a pivotal role in the pipeline. This stage involves aligning the reference image of a defect-free nameplate, stored locally, with a captured test image from the production line. The reference image and the captured image share the same layout, as both belong to the same customer. While the content of the engraved strings may differ between the two images, the logos remain identical, ensuring consistency in logo design. The alignment module ensures that the reference image is geometrically adjusted to match the captured image.

Initially, the reference and captured images are resized to the same dimensions to facilitate comparison and reduce inference time. Keypoints are detected in both images using a combination of ORB (Oriented FAST and Rotated BRIEF) and SIFT (Scale-Invariant Feature Transform) algorithms \cite{karami2017image,7827292}. ORB enables fast computation with rotational invariance, while SIFT enhances feature matching reliability by providing robustness against scale and rotation changes.

Feature descriptors from both images are matched using the Brute-Force Matcher \cite{BFMatcher}, and a ratio test is applied to retain the most relevant matches \cite{7827292}. The best matches obtained from ORB and SIFT are merged, and the top correspondences are selected for further processing. Using these selected feature correspondences, a homography matrix is computed with the RANSAC (Random Sample Consensus) algorithm \cite{zhang2008automatic,bazargani2018fast}. This matrix estimates the geometric transformation between the captured and reference images, accounting for rotation, scaling, and translation. The computed transformation is then applied to warp the reference image, ensuring alignment with the captured image.

Following alignment, logo coordinates are extracted from the aligned reference image using the YOLOv7 model, which is trained specifically for logo detection. A detailed explanation of YOLOv7 is provided in Section \ref{subsec: string detection and recognition}. The extracted coordinates from the aligned reference image are subsequently used to crop logos from the captured image, ensuring consistency in position and size. This step is essential for accurate defect detection.

\begin{figure}[h!]
  \centering
  \includegraphics[width=0.75\textwidth]{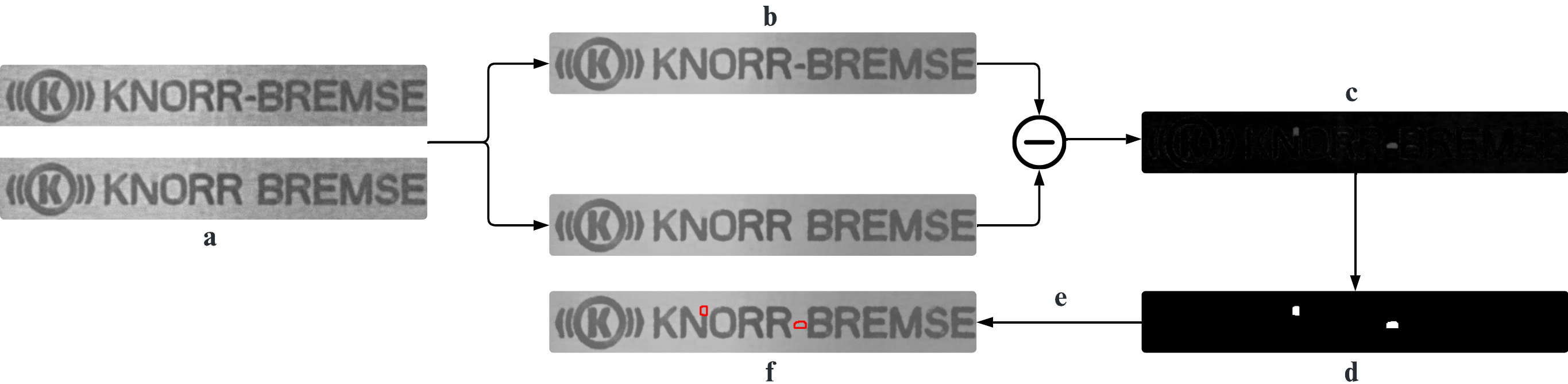}
  \caption{Logo defect detection.
    (a) Extracted reference (top) and captured images (bottom). 
    (b) Pre-processing with noise reduction and smoothing. 
    (c) Pixel-wise difference. 
    (d) Binary thresholding and morphological operations. 
    (e) Connected component analysis. 
    (f) Defect localization and visualization through contouring.}
  \label{fig:logo_defect_pipeline}
\end{figure}

\paragraph{Logo Defect Detection}

After alignment and logo extraction, defect detection is performed using traditional computer vision techniques, as illustrated in Figure \ref{fig:logo_defect_pipeline}. The extracted reference and captured logos first undergo noise reduction to minimize unwanted artifacts. A pixel-wise difference is computed between the two images to highlight discrepancies indicative of potential defects. The resulting difference image is then processed through binary thresholding to produce a binary map, accentuating defect regions.

Morphological operations, such as erosion and dilation \cite{Morphological}, are applied to the thresholded image to eliminate small noise while preserving significant defect regions. This step refines the detection process by reducing false positives and enhancing the visibility of larger anomalies. Connected component analysis \cite{connectedopencv} is then performed to identify contiguous regions corresponding to potential defects \cite{connectedcomp}. Only components within a defined area range are retained to filter out minor artifacts.

Finally, contours are drawn around the detected anomalies, effectively localizing defects on the captured image logo. This visualization highlights inconsistencies, offering a clear representation of defect location and severity, as shown in Figure \ref{fig:logo_defect_pipeline}(f). The combination of these steps ensures precise defect detection and localization, contributing to improved quality control in nameplate inspection.

\subsubsection{Experiments and Results}
\paragraph{Experimental Setup}

The rule-based computer vision approach does not require model training. However, to optimize defect detection, a grid search was performed to systematically identify the optimal thresholds for connected component analysis. These thresholds define the size range of connected components that are classified as significant defects. A range of minimum area threshold values were tested, and the optimal threshold was selected based on the performance metrics, including accuracy, precision, recall, and F1 score. This systematic approach ensured that the chosen thresholds maximized the detection performance.

For evaluation, a nameplate image with a resolution of $1920 \times 1600$ pixels, containing strings and logos, was used. A trained YOLOv7 model was employed to detect and extract logos from the nameplate. The method was assessed on a dataset comprising 345 logo images, categorized into two classes: 195 non-defective (good) logos and 150 defective (bad) logos. Logo sizes varied due to the presence of different logo types on the nameplate. Images were captured under controlled conditions, but variations in lighting to evaluate the robustness of the method.

\paragraph{Results}

The performance of the method was evaluated for the classification of defective and non-defective nameplate content using common metrics, including accuracy, precision, recall, and F1 score. Although the method localizes defects, localization performance was not assessed due to the absence of segmented ground truth masks. The evaluation yielded an accuracy of 97.9\%, a precision of 95.5\%, a recall of 100\%, and an F1 score of 97.7\%. Seven non-defective logos were misclassified as defective, resulting in a slight reduction in precision.

\paragraph{Discussion}

The method demonstrates high accuracy in detecting defective logos under stable lighting conditions. However, performance is affected by lighting variations, such as overexposure or underexposure, which introduce noise and influence the difference calculation. The fixed threshold values applied during the binary thresholding stage may lead to false positives under varying lighting conditions. In some cases, surface scratches, which are not classified as defects in the current use case, were mistakenly detected as anomalies, figure \ref{fig:logo_scracth}

To enhance performance across different lighting environments, adaptive thresholding techniques or illumination normalization methods could be introduced. Additionally, more robust pre-processing steps to account for lighting inconsistencies could further improve the reliability of the method. However, since the current approach relies on pixel-wise difference calculations, it does not inherently differentiate between actual defects and surface scratches. As a result, any variations on the surface, including scratches, may be incorrectly flagged as defects. This represents a fundamental limitation of the method, as surface irregularities that do not affect the functionality of the nameplate should ideally be ignored. 

\subsection{Detection and Recognition}
\label{subsec: string detection and recognition}
\subsubsection{Models and Process}

The objective of this process is to detect strings and logos on laser-engraved nameplates using the YOLOv7 \cite{wang2023yolov7} model, while Tesseract OCR \cite{smith2007overview} is employed for character detection and recognition.

\paragraph{String and Logo Detection}

YOLOv7 is a deep learning model recognized for its high detection speed and accuracy across its various configurations. One such configuration is YOLOv7-tiny, which features a reduced network depth and width, enabling faster detection while maintaining competitive accuracy. This balance makes YOLOv7-tiny particularly well-suited for tasks involving industrial nameplate inspection, where rapid detection is required without compromising precision.

In this paper, YOLOv7-tiny was selected to detect strings and logos. During inference, the YOLO model receives two images from the alignment module, which include an aligned reference image and a captured image. The model processes both images to extract strings and logos, generating their bounding-box coordinates. Logos extracted from the reference image, along with their bounding-box coordinates, are passed to the logo defect detection module. The strings of the captured image are forwarded to the Tesseract OCR model for further processing and recognition. This two-step approach ensures that logos undergo defect detection, while strings proceed to the character detection and recognition stage, allowing discrepancies between the detected and expected values from the MES to be identified.

\begin{figure}[h!]
  \centering
  \includegraphics[width=0.75\textwidth]{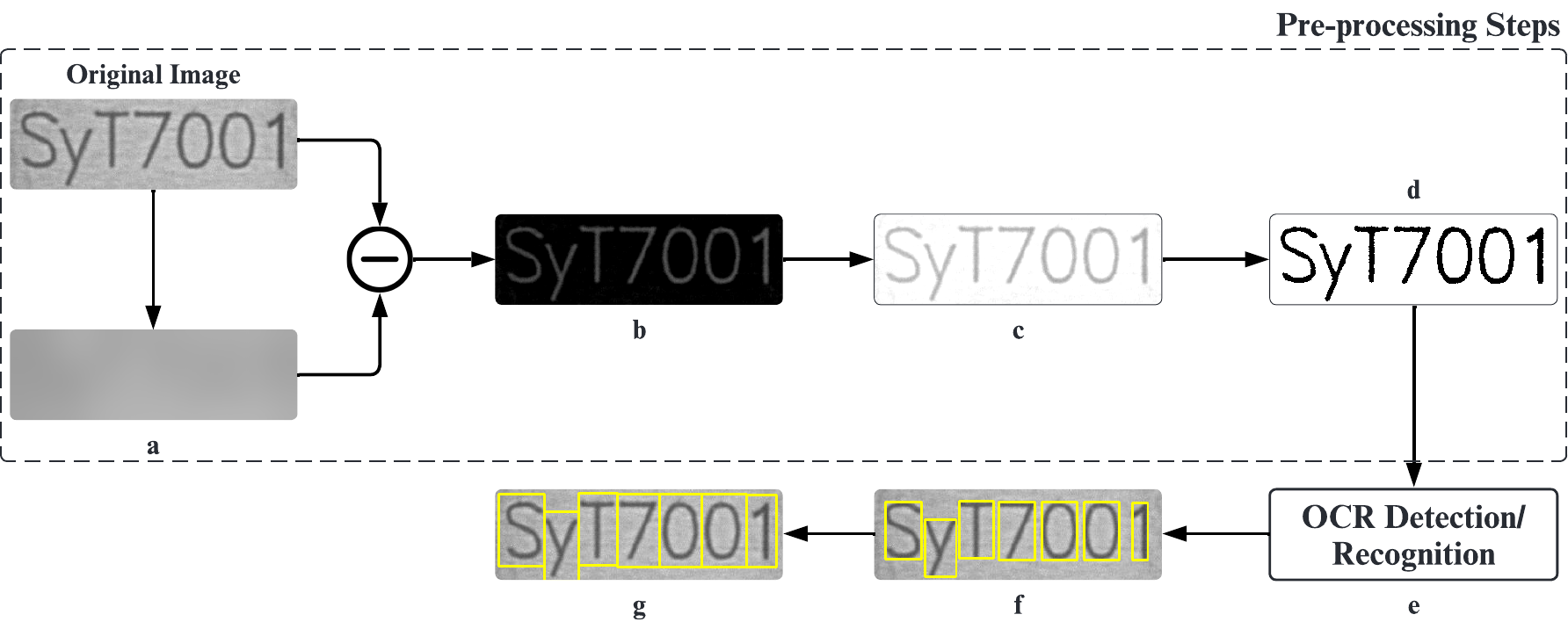}
  \caption{Character Detection and Recognition. (a) Gaussian Blur, (b) Background subtraction, (c) Bitwise NOT, 
    (d) Binary thresholding and morphological operations, 
    (e) Tesseract OCR detection/recognition, 
    (f) OCR output, (g) Bounding Box Adjustment}
  \label{fig:Character Detection and Recognition}
\end{figure}

\paragraph{Character detection and Recognition}

For character recognition, Tesseract 4 \cite{gittess}, configured with the Long-Short-Term Memory (LSTM) engine, was selected to improve accuracy and processing speed. Adjustments to the Page Segmentation Modes (PSM) were made to improve text detection accuracy. Based on experimental evaluations and Tesseract documentation \cite{tessgit}, PSM 7 was chosen due to its optimization for single-line text detection, which aligns with the requirements of this project.

Before passing the strings to Tesseract OCR for recognition, several pre-processing steps were applied to improve detection accuracy. The pre-processing workflow, illustrated in Figure \ref{fig:Character Detection and Recognition}, outlines the sequence of operations involved in preparing the input images. The string images extracted from the YOLO-detected regions of the captured nameplates were first processed with Gaussian blur to estimate and smooth the background. A kernel size of \(149 \times 149\) was chosen to suppress finer details and local variations, effectively capturing the overall background intensity. The background was then subtracted from the original image to isolate the text region. Subsequently, the image was inverted, resulting in black text on a white background. This inversion is a requirement for Tesseract OCR, as it operates on images with this format.

To enhance the clarity of string content, binary thresholding was applied to binarize the image, ensuring a distinct separation between the foreground and background. Morphological closing operations \cite{Morphological} were subsequently performed to fill small gaps, further refining the binary image and preserving the integrity of detected characters. These pre-processing steps, as shown in Figure \ref{fig:Character Detection and Recognition}, ensure the string content is well-defined before being passed to Tesseract OCR for character detection and recognition.

Bounding box coordinates generated by Tesseract OCR were evaluated to verify accurate character localization. In cases where the bounding boxes appeared closely packed or overlapping, an additional step was applied to adjust the bounding boxes by adding padding to the coordinates. The adjusted bounding boxes, shown in Figure \ref{fig:Character Detection and Recognition}(g), ensured sufficient spacing between characters, minimizing the risk of errors and enhancing the reliability of the OCR output.

\subsubsection{Experiments and Results}
\paragraph{Experimental Setup}

The experimental setup is divided into two parts to evaluate the performance of the YOLOv7-tiny model and the Tesseract OCR model.

\paragraph{YOLOv7-tiny}

The dataset used for training the YOLOv7-tiny model consists of nameplate images from the production line, each with a resolution of $1920 \times 1600$ pixels. The dataset exclusively contains defect-free logos and strings. A total of 476 nameplate images were collected and used for training. Before training, each input image was resized to $640 \times 640$ pixels. The model was trained to classify strings, logos, and Data Matrix Codes (DMC) as individual classes.

The training process involved 80 epochs, with a batch size of 64. The momentum was set to 0.937, the initial learning rate was set to 0.01, and the weight decay for learning rate adjustment was 0.0005 \cite{yolov7_hyp}.

To evaluate the YOLOv7 model, a separate set of 65 non-defective nameplate images was selected. Each image contained logos, strings, and DMCs, and these images were captured under controlled lighting conditions, along with a subset taken under varying lighting conditions to test robustness.

\paragraph{Tesseract OCR}

For the development and evaluation of the OCR pipeline, the \textbf{tessdata-best} pre-trained Tesseract OCR model \cite{tessmodel} was utilized. Since this model had already been trained on English-language documents and demonstrated superior accuracy compared to other available models, no further fine-tuning was performed during development. The same dataset of 65 nameplate images used for YOLO evaluation was employed for assessing the performance of Tesseract OCR. The extracted strings were compared against the ground truth to evaluate the accuracy of the OCR system.

\paragraph{Results}

The performance of the YOLOv7-tiny model was assessed using standard evaluation metrics, including accuracy, precision, recall, and F1 score. Additionally, two key metrics—mean average precision (mAP) at IoU 0.5 (mAP@.5) and mAP at IoU thresholds ranging from 0.5 to 0.95 (mAP@.5:.95)—were used to provide a comprehensive evaluation. The YOLOv7-tiny model achieved 100\% for accuracy, precision, recall, and F1 score. The mAP@.5 also reported 100\%, while mAP@.5:.95 achieved 98.9\%, indicating the model’s high capability to detect all classes without generating false positives.

The evaluation of Tesseract OCR was conducted using the Natural Language Toolkit (NLTK) library, specifically leveraging the edit distance calculation to compare OCR predictions with ground truth text \cite{nltk}. This method quantifies discrepancies by measuring the minimum number of character edits required to transform the OCR output into the correct string.

Across the 65 nameplate images, a total of 390 strings were detected. Tesseract OCR misclassified five strings, resulting in a word-level accuracy of 98.71\% and a word error rate (WER) of 1.28\%. Additionally, character-level accuracy was evaluated since Tesseract OCR operates at the character level. Out of 2,600 total characters, six incorrect insertions were recorded, yielding a character-level accuracy of 99.79\% and a character error rate (CER) of 0.21\%.

\paragraph{Discussion}

The YOLOv7-tiny model consistently demonstrated flawless detection across all test cases, correctly identifying logos, strings, and DMCs without any false positives or misclassifications. However, the Tesseract OCR model exhibited occasional mispredictions by inserting extra characters into the detected strings. This occurred even in cases where the images were free from noise or artifacts, and there was clear separation between the foreground and background. This behavior highlights a limitation of the Tesseract OCR engine, where excessive sensitivity to minor variations can lead to over-segmentation and character insertion errors.

Despite these minor inaccuracies, the overall performance of the OCR system remains robust, achieving high accuracy at both word and character levels. Future improvements could involve integrating additional pre-processing techniques or exploring adaptive models to mitigate errors associated with character insertion.

\subsection{Anomaly Detection on Single Characters}
\label{subsec: anomaly detection on single characters}
In the inspection pipeline, ensuring the quality of nameplates requires detecting and localizing defects at the character level. Examples of such defective characters are shown in Figures \ref{fig:Original image with defect}, \ref{fig:original_resized}. As detailed in Section \ref{subsec: string detection and recognition}, individual characters were extracted from nameplates using a combination of models. YOLO was utilized for string detection, and Tesseract OCR was employed to isolate and extract individual characters from the detected strings. This pre-processing step was necessary to simplify the complexity of the problem by focusing on character-level reconstruction rather than processing entire nameplates.

The decision to adopt an unsupervised approach for anomaly detection was driven by the limited availability of defective data. Supervised methods, such as object detection models, require extensive datasets containing labeled defective samples, which were not feasible to obtain due to the rarity of defective nameplates in production. Instead, an unsupervised method based on generative modeling was employed. Variational autoencoders (VAEs), which are generative models by design, were selected for this task.

Training a model to reconstruct entire nameplates presented challenges due to the variability in string content across different nameplates. Nameplates contain strings that differ in length, character arrangement, and content, making it difficult to capture all possible variations during training. This variability hindered the model’s ability to generalize and reconstruct nameplates accurately. To address this, the focus was shifted to training the model on individual characters, significantly reducing variability and enabling more precise defect localization. This character-based approach simplified the reconstruction task while ensuring higher-quality outputs.

Initially, a traditional autoencoder was considered for the task. However, traditional autoencoders often suffer from overfitting and struggle to generalize to unseen data, particularly in unsupervised anomaly detection \cite{kingma2013auto, chen2023auto}. To address these limitations, a Variational Autoencoder (VAE) was introduced \cite{kingma2013auto, kingma2019introduction}. The probabilistic nature of VAEs enables better generalization, making them more robust for character-level reconstruction tasks.

\begin{figure}[h!]
  \centering
  \includegraphics[width=0.95\textwidth]{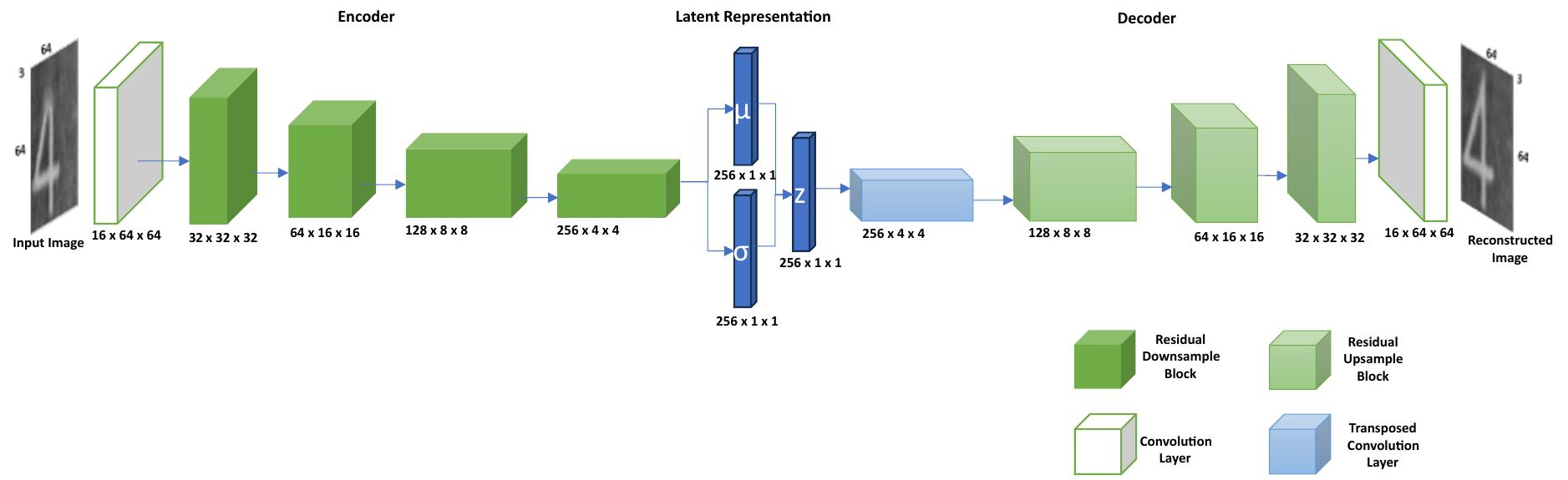}
  \caption{Block architecture of the proposed Residual Variational Autoencoder.}
  \label{fig:your-image-label}
\end{figure}

To further improve reconstruction quality and address issues such as vanishing gradients, residual blocks were incorporated into the VAE architecture. Residual connections facilitate the flow of gradients during backpropagation by allowing certain layers to be bypassed, making the model easier to train and enhancing its ability to learn complex patterns \cite{borawar2023resnet}. The enhanced architecture, shown in Figure \ref{fig:your-image-label}, was specifically designed to reconstruct defective characters into their non-defective ones with high fidelity.

After reconstructing the characters, post-processing methods were applied to localize and classify anomalies. These methods ensure that defects are identified and localized at the character level with precision, offering a practical and effective solution to the challenge of detecting defects on nameplates with limited defective data.

\subsubsection{Architecture of Residual Variational Autoencoder (ResVAE)}

The Residual Variational Autoencoder (ResVAE) \cite{10104785, appati2023deep}  follows a structure similar to the standard Variational Autoencoder (VAE), comprising an encoder, a decoder, and a latent space. In ResVAE, the encoder network maps the input data \( x \) to a latent representation through a series of residual downsampling blocks, while the decoder reconstructs the data from the latent representation using residual upsampling blocks. The overall block architecture of the ResVAE is illustrated in Figure \ref{fig:your-image-label}, depicting the sequence of layers involved in both the encoder and decoder pathways.

The encoder processes the input data by first applying a convolutional layer, which increases the number of channels while preserving the spatial resolution of the input image. As shown in Figure \ref{fig:your-image-label}, this initial convolution is followed by four sequential residual downsampling blocks, referred to as ResDown. Each block progressively reduces the spatial dimensions by a factor of two and doubles the number of feature channels, resulting in a compact latent representation that encodes the essential characteristics of the input image, as described in the table \ref{tab:resvae-architecture}. At the final stage of the encoder, two parallel convolutional layers compute the latent distribution parameters, specifically the mean vector \( \boldsymbol{\mu} \) and the logarithm of the variance \( \log \boldsymbol{\sigma}^2 \), which together define a multivariate Gaussian distribution in the latent space.

During training, the latent vector \( \boldsymbol{z} \) is sampled from this distribution by applying the reparameterization trick:

\begin{equation}
\boldsymbol{z} = \boldsymbol{\mu} + \boldsymbol{\epsilon} \odot \exp\left(0.5 \log \boldsymbol{\sigma}^2\right),
\label{eq:reparameterization}
\end{equation}

where \( \boldsymbol{\epsilon} \sim \mathcal{N}(0, \mathbf{I}) \) is a random variable drawn from a standard normal distribution, and \( \odot \) denotes element-wise multiplication. This technique ensures that the sampling process remains differentiable, enabling backpropagation to optimize the network parameters.

The decoder mirrors the encoder’s structure, beginning with a transposed convolution that upscales the latent vector \( \boldsymbol{z} \). As depicted in Figure \ref{fig:your-image-label}, the decoder consists of three residual upsampling blocks, termed ResUp, which sequentially double the spatial dimensions while reducing the number of channels. The residual connections in these blocks play a crucial role in preserving fine-grained details during reconstruction. The final output layer applies a \(3 \times 3\) convolution followed by a hyperbolic tangent activation function \cite{6026515}, scaling the pixel values to the range \([-1, 1]\) to generate the reconstructed image \( \hat{x} \).

\begin{table}[htbp]
\centering
\caption{Model Architecture for ResVAE – Detailed Breakdown of a Single Residual Downsample and Upsample Block}
\renewcommand{\arraystretch}{1.2} 
\small 
\begin{tabular}{|l|l|l|l|l|l|}
\hline
& \textbf{Layer} & \textbf{Input Dim} & \textbf{Operation} & \textbf{Kernel/Stride} & \textbf{Output Dim} \\
\hline
\multirow{6}{*}{Encoder} & Input Layer & 3 x 64 x 64  & Conv2d & 7x7 / 1 & 16 x 64 x 64 \\
\cline{2-6}
& \multicolumn{5}{c|}{\textbf{ResDown Block}} \\
\cline{2-6}
& Conv2d (1st) & 16 x 64 x 64  & Conv2d & 3x3 / 2 & 32 x 32 x 32 \\
\cline{2-6}
& Conv2d (2nd) & 32 x 32 x 32  & Conv2d & 3x3 / 1 & 32 x 32 x 32 \\
\cline{2-6}
& Skip Connection & 16 x 64 x 64  & Conv2d & 3x3 / 2 & 32 x 32 x 32 \\
\cline{2-6}
& Residual Addition & 32 x 32 x 32  & Element-wise Add & N/A & 32 x 32 x 32 \\
\hline
\multirow{2}{*}{Latent Space} & Conv2d ($\mu$, log($\sigma^2$)) & 256 x 4 x 4 & Conv2d & 4x4 / 1 & 256 x 1 x 1 \\
\cline{2-6}
& Latent Vector (z) & 256 x 1 x 1 & Sampling & N/A & 256 x 1 x 1 \\
\hline
\multirow{6}{*}{Decoder} & Upsample Layer & 256 x 1 x 1 & ConvTranspose2d & 4x4 / 1 & 256 x 4 x 4 \\
\cline{2-6}
& \multicolumn{5}{c|}{\textbf{ResUp Block}} \\
\cline{2-6}
& Upsample Layer & 256 x 4 x 4 & Nearest Neighbor Interpolation & Scale Factor = 2 & 256 x 8 x 8 \\
\cline{2-6}
& Conv2d (1st) & 256 x 8 x 8 & Conv2d & 3x3 / 1 & 128 x 8 x 8 \\
\cline{2-6}
& Conv2d (2nd) & 128 x 8 x 8 & Conv2d & 3x3 / 1 & 128 x 8 x 8 \\
\cline{2-6}
& Skip Connection & 256 x 8 x 8 & Conv2d & 3x3 / 1 & 128 x 8 x 8 \\
\cline{2-6}
& Residual Addition & 128 x 8 x 8 & Element-wise Add & N/A & 128 x 8 x 8 \\
\hline
& Output Layer & 128 x 8 x 8 & Conv2d (Tanh) & 3x3 / 1 & 3 x 64 x 64 \\
\hline
\end{tabular}
\label{tab:resvae-architecture}
\end{table}

\subsubsection{Structure of Residual Block}

In the downsampling residual block, the main path begins with a convolutional layer that reduces the spatial dimensions by applying a stride of two, refer to table \ref{tab:resvae-architecture} for the implementation. This operation is followed by batch normalization and activation using the ELU function, enhancing nonlinearity. A second convolutional layer further processes the features without additional spatial reduction. Simultaneously, the shortcut connection applies a single convolutional layer with matching stride and kernel size to project the input directly into the reduced feature space. Both pathways ensure output tensors of identical dimensions, allowing their element-wise addition. This residual connection preserves key features from the input while enabling the block to efficiently reduce spatial resolution and enhance feature representation.

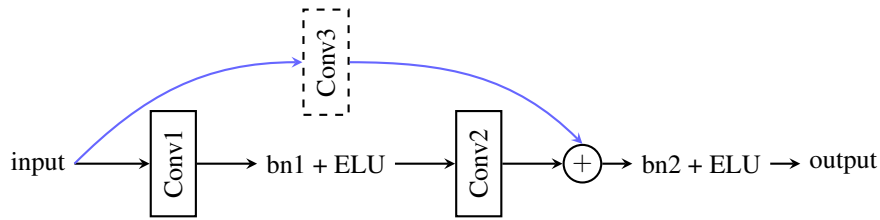
\begin{figure}[h!]
    \centering
    \begin{tikzpicture}[
        scale=0.9, 
        box/.style={draw, minimum width=0.6cm, minimum height=1.4cm, thick}, 
        dashbox/.style={draw, minimum width=0.6cm, minimum height=1.4cm, thick, dashed}, 
        arrow/.style={->,>=stealth, thick},
        skip/.style={->,>=stealth, thick, blue!60},
        plus/.style={circle, draw, minimum size=0.5cm, inner sep=0pt, thick}, 
        node distance=1.5cm 
    ]
    \node (input) {input};
    \node[box, right=1cm of input] (conv1) {};
    \node at (conv1.center) {\rotatebox{90}{Conv1}};
    \node[right=0.8cm of conv1] (elu1) {bn1 + ELU};
    \node[box, right=0.8cm of elu1] (conv2) {};
    \node at (conv2.center) {\rotatebox{90}{Conv2}};
    \node[dashbox] (conv3) at ($(conv1)!0.5!(conv2)+(0,1.5cm)$) {};
    \node at (conv3.center) {\rotatebox{90}{Conv3}};
    \node[plus, right=0.8cm of conv2] (add) {$+$};
    \node[right=0.4cm of add] (elu2) {bn2 + ELU};
    \node[right=0.4cm of elu2] (output) {output};
    \draw[arrow] (input) -- (conv1);
    \draw[arrow] (conv1) -- (elu1);
    \draw[arrow] (elu1) -- (conv2);
    \draw[arrow] (conv2) -- (add);
    \draw[arrow] (add) -- (elu2);
    \draw[arrow] (elu2) -- (output);
    \draw[skip] (input.east) to[out=45, in=180] (conv3.west);
    \draw[skip] (conv3.east) to[out=0, in=135] (add.north);
    \end{tikzpicture}

    \caption{Structure of the Proposed Residual Block.}
    \label{fig:residual_block}
\end{figure}

The upsampling residual block mirrors this structure, adapting it for spatial enlargement. The main path begins by applying nearest-neighbor interpolation to increase spatial dimensions. The enlarged feature maps are processed through a sequence of two convolutional layers, interspersed with batch normalization and ELU activation, to refine and restore spatial details. The shortcut connection projects the interpolated input directly into the enlarged feature space through a single convolutional layer. The outputs of the two pathways are added element-wise, creating a residual connection that maintains the integrity of the input features while ensuring the spatial dimensions are accurately restored.

\subsubsection{Loss Function}
\label{loss function}

During training, the ResVAE model learns to encode the input \( x \) into a latent space parameterized by the mean \( \boldsymbol{\mu} \) and standard deviation \( \boldsymbol{\sigma} \) of a Gaussian distribution. The latent vector \( \boldsymbol{z} \) is sampled using the reparameterization trick (as described in Equation~\ref{eq:reparameterization}) from this distribution. To achieve effective training, the total loss function is composed of several terms that optimize the reconstruction quality while regularizing the structure of the latent space. The primary loss functions used are the Kullback-Leibler (KL) divergence loss, Mean Squared Error (MSE) reconstruction loss, Structural Similarity Index Measure (SSIM) loss, and Feature Perceptual loss. Each of these is discussed below.

\paragraph{Kullback-Leibler (KL) Divergence Loss}

The KL divergence loss measures the difference between the approximate posterior distribution \( q(\boldsymbol{z} \mid x) \) and a standard Gaussian prior distribution \( p(\boldsymbol{z}) = \mathcal{N}(0, \mathbf{I}) \). This term acts as a regularizer, ensuring that the learned latent space remains close to the standard Gaussian prior. By enforcing this regularization, the latent space is encouraged to be smooth and well-behaved, allowing for more effective generation and interpolation of data points.

The KL divergence loss is given by:

\begin{equation}
\mathcal{L}_{\text{KL}} = -\frac{1}{2} \sum_{j=1}^{d} \left(1 + \log \sigma_j^2 - \mu_j^2 - \sigma_j^2\right),
\label{eq:kl_divergence}
\end{equation}

where \( \boldsymbol{\mu} = [\mu_1, \mu_2, \dots, \mu_d] \) and \( \boldsymbol{\sigma}^2 = [\sigma_1^2, \sigma_2^2, \dots, \sigma_d^2] \) are the mean and variance vectors for each latent dimension \( j \), and \( d \) is the dimensionality of the latent space.

\paragraph{Mean Squared Error (MSE) Loss}

The reconstruction loss, often expressed as the Mean Squared Error (MSE), quantifies the difference between the original input \( x \) and the reconstructed output \( \hat{x} \). The MSE loss penalizes large pixel-wise differences, ensuring that the reconstructed image closely resembles the original.

The MSE loss is given by:

\begin{equation}
\mathcal{L}_{\text{MSE}} = \frac{1}{n} \sum_{i=1}^{n} \left( x_i - \hat{x}_i \right)^2,
\label{eq:mse_loss}
\end{equation}

where \( n \) is the number of pixels, and \( x_i \) and \( \hat{x}_i \) represent the pixel values of the original and reconstructed images, respectively. This loss ensures that the model minimizes pixel-level discrepancies between the input and its reconstruction. However, MSE alone may fail to capture perceptual similarities between the images.

\paragraph{Structural Similarity Index Measure (SSIM) Loss}

While the MSE loss is effective for measuring pixel-level differences, it does not account for the perceptual quality of the reconstructed image \cite{9345813}. To address this, the Structural Similarity Index Measure (SSIM) \cite{1284395, nilsson2020understanding} loss is introduced, which compares the input and reconstructed images based on three perceptual aspects: luminance, contrast, and structure.

The SSIM between images \( x \) and \( \hat{x} \) is defined as:

\begin{equation}
\text{SSIM}(x, \hat{x}) = \frac{ (2 \mu_x \mu_{\hat{x}} + C_1) (2 \sigma_{x \hat{x}} + C_2) }{ (\mu_x^2 + \mu_{\hat{x}}^2 + C_1) (\sigma_x^2 + \sigma_{\hat{x}}^2 + C_2) },
\label{eq:ssim}
\end{equation}

where \( \mu_x \) and \( \mu_{\hat{x}} \) are the means of \( x \) and \( \hat{x} \), \( \sigma_x^2 \) and \( \sigma_{\hat{x}}^2 \) are the variances, \( \sigma_{x \hat{x}} \) is the covariance between \( x \) and \( \hat{x} \), and \( C_1 \) and \( C_2 \) are small constants to stabilize the division when the denominators are close to zero.

The SSIM loss is computed as:

\begin{equation}
\mathcal{L}_{\text{SSIM}} = 1 - \text{SSIM}(x, \hat{x}),
\label{eq:ssim_loss}
\end{equation}

so that minimizing the loss corresponds to maximizing the structural similarity between the original and reconstructed images \cite{zhao2016loss}. This loss focuses on structural features, making it more aligned with human visual perception than MSE alone.

\paragraph{Feature Perceptual Loss}
\label{para:Feature Perceptual Loss}

To further improve the perceptual quality of the reconstructed images, a feature perceptual loss \cite{johnson2016perceptual} is employed. Unlike pixel-wise losses such as MSE, which focus on individual pixel differences, the perceptual loss compares high-level features extracted from a pre-trained network to evaluate the similarity between the original and reconstructed images \cite{7926714}.

The feature perceptual loss is computed by passing both the original and reconstructed images through a series of convolutional layers initialized with pre-trained weights from the VGG19 network. The loss is defined as the Mean Squared Error between the feature maps at multiple layers:

\begin{equation}
\mathcal{L}_{\text{perceptual}} = \frac{1}{L} \sum_{l=1}^{L} \frac{1}{n_l} \sum_{i=1}^{n_l} \left( \phi_l(x)_i - \phi_l(\hat{x})_i \right)^2,
\label{eq:perceptual_loss}
\end{equation}

where \( L \) is the number of layers used, \( n_l \) is the number of features at layer \( l \), and \( \phi_l(x) \) and \( \phi_l(\hat{x}) \) are the feature representations of the original and reconstructed images at layer \( l \). By focusing on high-level features rather than individual pixels, this loss captures perceptual differences that are more meaningful for visual quality, helping to reduce artifacts and improve the realism of the reconstructed images.

In this implementation, a custom network is manually defined to replicate the convolutional layers of the VGG19 architecture \cite{vgg19, simonyan2014very}, pre-trained on the ImageNet-1k dataset. The network contains 16 convolutional layers with filter sizes increasing from 64 to 512. Each convolution employs a 3$\times$3 kernel with stride 1 and padding 1 to maintain spatial dimensions. Max-pooling layers with a 2$\times$2 kernel and stride 2 are placed between groups of layers, progressively reducing the spatial resolution while capturing features at multiple scales. This design mostly matches VGG19, but the final pooling layer is omitted, leading to four pooling stages instead of five, and the first ReLU is placed in a slightly different location. Despite these small deviations, the core feature extraction process remains similar to the original VGG19 backbone.

The weights and biases of these convolutional layers are initialized using pre-trained VGG19 parameters, allowing the network to leverage features learned from ImageNet. These parameters enable the convolutional layers to detect low-level features such as edges and textures, as well as high-level semantic patterns such as shapes and object parts. The fully connected layers of the original VGG19 architecture are excluded, as they are specific to classification tasks. By focusing solely on convolutional layers, the network extracts spatially localized and semantically meaningful features that are used to compute the feature perceptual loss.

\paragraph{Total Loss Function}

The total loss function \(\mathcal{L}_{\text{total}}\) used to train the ResVAE model is a weighted combination of multiple loss terms and is defined as:

\begin{equation}
\mathcal{L}_{\text{total}} = \alpha \, \mathcal{L}_{\text{MSE}} + \beta \, \mathcal{L}_{\text{KL}} + \gamma \, \mathcal{L}_{\text{SSIM}} + \kappa \, \mathcal{L}_{\text{perceptual}},
\label{eq:total_loss}
\end{equation}

where \(\alpha\), \(\beta\), \(\gamma\), and \(\kappa\) are hyperparameters that control the relative importance of each loss term during training. These weights are crucial for balancing the influence of different losses to ensure that the model learns meaningful latent representations while producing high-quality reconstructions. 

Each loss component serves a specific role in the training process. The Mean Squared Error (MSE) loss \(\mathcal{L}_{\text{MSE}}\) captures pixel-level discrepancies between the original input \(x\) and the reconstructed output \(\hat{x}\), providing a straightforward measure of reconstruction error. The Kullback-Leibler (KL) divergence loss \(\mathcal{L}_{\text{KL}}\) acts as a regularizer for the latent space, enforcing the posterior distribution \(q(\boldsymbol{z} \mid x)\) to be close to the prior \(p(\boldsymbol{z}) = \mathcal{N}(0, \mathbf{I})\). This regularization encourages a smooth and well-behaved latent space, facilitating effective generation and interpolation of data points. The Structural Similarity Index Measure (SSIM) loss \(\mathcal{L}_{\text{SSIM}}\) focuses on preserving perceptual fidelity by encouraging the model to maintain structural coherence between the original and reconstructed images. Lastly, the perceptual loss \(\mathcal{L}_{\text{perceptual}}\) ensures that the reconstructed images retain meaningful feature representations by comparing feature maps extracted at multiple levels of abstraction from a pre-trained network, aligning the reconstructions more closely with human visual perception.

Incorporating multiple loss functions presents the challenge of ensuring proper balance between them. If any loss term dominates the optimization process, it can lead to overfitting specific aspects of the data. For example, if the weight on the MSE loss (\(\alpha\)) is too large, the model may prioritize minimizing pixel-wise errors at the expense of perceptual quality. Conversely, if the KL divergence weight (\(\beta\)) is overemphasized, the model may fail to adequately reconstruct the input, as it focuses too heavily on regularizing the latent space. In such cases, imbalanced loss terms can cause the gradients from one loss to overshadow others, leading to sub-optimal convergence behavior and the potential for the model to become trapped in poor local minima. Moreover, the inclusion of multiple loss terms inherently increases computational complexity. If the loss weights are not appropriately tuned, this additional computational cost may lead to inefficient learning, as the model might oscillate between different objectives during optimization, resulting in slower and less stable convergence.

To address these issues, the weights for each loss term were determined empirically through a series of validation experiments. By adjusting the values of \(\alpha\), \(\beta\), \(\gamma\), and \(\kappa\) based on the model's observed performance on a validation set, the tuning process aimed to improve convergence and achieve a reasonable trade-off between reconstruction accuracy and training efficiency. Although this empirical approach does not guarantee optimal results, it facilitated practical convergence during training and helped balance the competing objectives of the loss terms.

In this work, multiple experiments were conducted to evaluate the impact of each loss term and the effect of tuning their weights. These experiments, discussed in subsequent sections, demonstrate the importance of each loss component and the role of loss balancing in achieving high-quality reconstructions and meaningful latent representations.

\subsubsection{Implementation and Training}

\paragraph{Implementation}  
The Residual Variational Autoencoder (ResVAE) is designed to efficiently encode and decode input images of size \(64 \times 64 \times 3\). As shown in Table~\ref{tab:resvae-architecture}, the encoder consists of an initial convolutional layer, followed by a series of residual downsampling blocks (ResDown) that progressively reduce spatial dimensions while increasing channel depth. The latent representation is obtained through two convolutional layers, encoding the mean \( \boldsymbol{\mu} \) and the logarithm of the variance \( \log \boldsymbol{\sigma}^2 \).  

The decoder mirrors this structure, employing upsampling residual blocks (ResUp) to gradually reconstruct the spatial dimensions while reducing the channel depth. A final convolutional layer with a hyperbolic tangent activation function generates the reconstructed image. The full architectural details, including kernel sizes, operations, and output dimensions, are provided in Table~\ref{tab:resvae-architecture}.

Increasing the number of channels in the initial convolutional layers and expanding the latent space can enhance reconstruction quality by allowing the model to learn and represent more features. However, increasing the number of channels and the latent space size also leads to a larger model with more parameters, resulting in higher computational costs and longer inference times.  

\paragraph{Dataset}  

The ResVAE model was trained using a dataset comprising 2,957 images of individual characters extracted from various nameplates. These characters include ten digits (0–9) and seven letters (S, C, B, K, Y, T, A), which were specific to the nameplates under consideration. To increase the size and diversity of the dataset, augmentations such as scaling, translation, rotation, and flipping were applied, resulting in a final dataset of 6,901 images.

Two datasets were prepared before training, each containing these 6,901 images, but with distinct purposes. The first dataset primarily consisted of non-defective images, supplemented by a smaller subset of defective images. To create defective samples, 10 images for each character (a total of 170 images) were manually manipulated using the GIMP tool \cite{cutts1997introduction} to introduce realistic defects. Augmentations, including scaling, translation, and rotation, were applied to both defective and non-defective samples to improve the model's robustness to variations.

The second dataset, serving as the ground truth, consisted exclusively of non-defective images, providing a clean reference for evaluating reconstruction quality. Identical augmentations were applied to this dataset to ensure consistency with the first dataset. Using two datasets during training allowed the model to learn how defective characters appear and to attempt to reconstruct non-defective versions. This dual-dataset approach improved the model’s ability to distinguish between defective and non-defective characters.

For validation, two separate datasets were prepared, each containing 206 images, with the same dual-structure setup as the training datasets. These validation datasets were used to monitor the model's performance during training, ensuring that it could accurately reconstruct non-defective characters while learning to recognize defects.

\begin{figure}[htbp]
    \centering
    \begin{subfigure}{0.15\textwidth}
        \includegraphics[width=\linewidth]{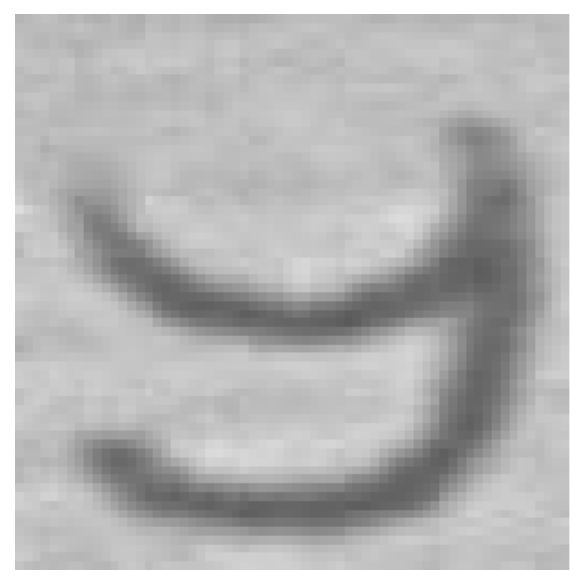}
        \caption{}
        \label{fig:original_resized}
    \end{subfigure} \hfill
    \begin{subfigure}{0.15\textwidth}
        \includegraphics[width=\linewidth]{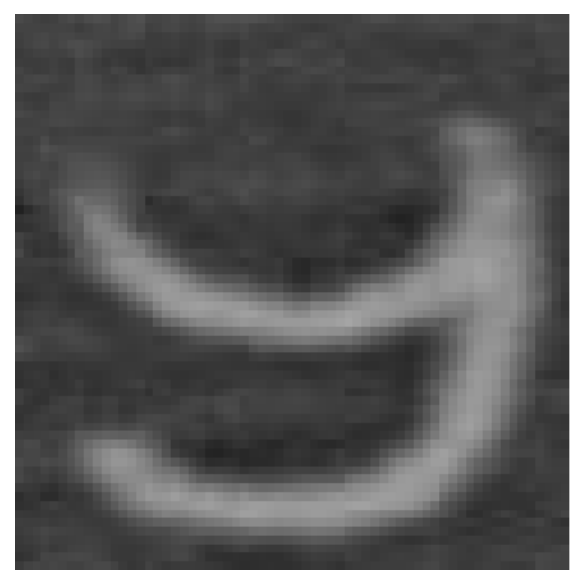}
        \caption{}
        \label{fig:original_inverted}
    \end{subfigure} \hfill
    \begin{subfigure}{0.15\textwidth}
        \includegraphics[width=\linewidth]{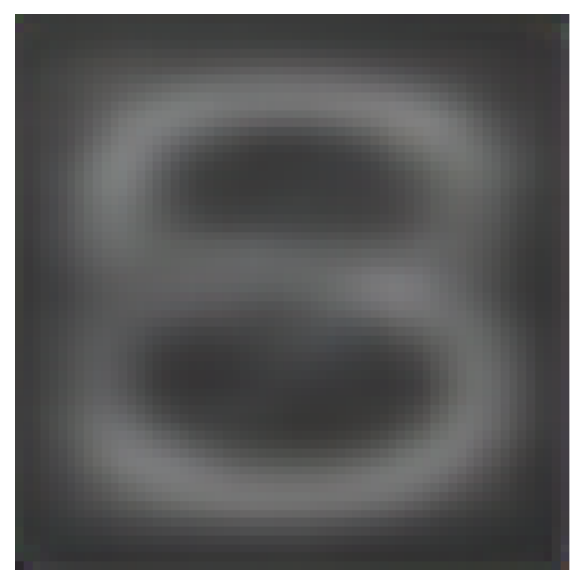}
        \caption{}
        \label{fig:reconstruct_mse}
    \end{subfigure} \hfill
    \begin{subfigure}{0.15\textwidth}
        \includegraphics[width=\linewidth]{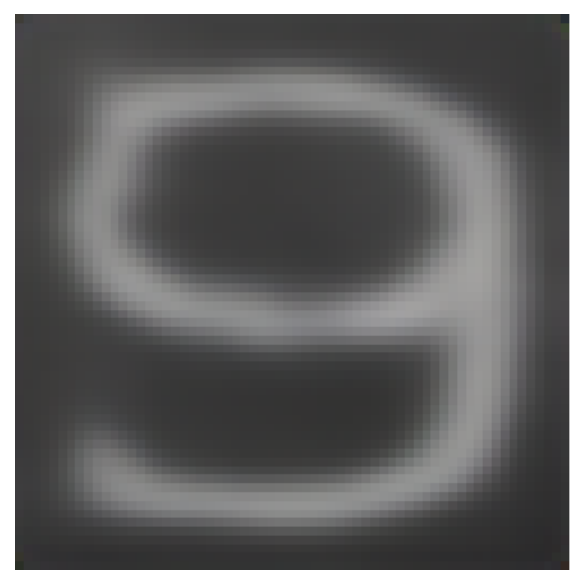}
        \caption{}
        \label{fig:reconstruct_mse_ssim}
    \end{subfigure} \hfill
    \begin{subfigure}{0.15\textwidth}
        \includegraphics[width=\linewidth]{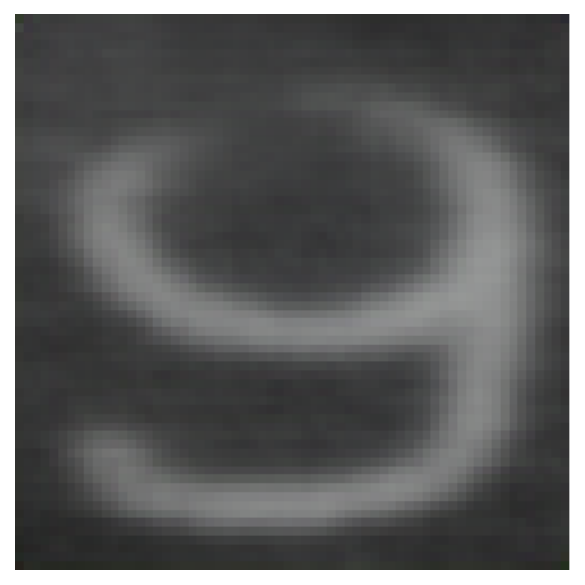}
        \caption{}
        \label{fig:reconstruct_mse_feature}
    \end{subfigure} \hfill
    \begin{subfigure}{0.15\textwidth}
        \includegraphics[width=\linewidth]{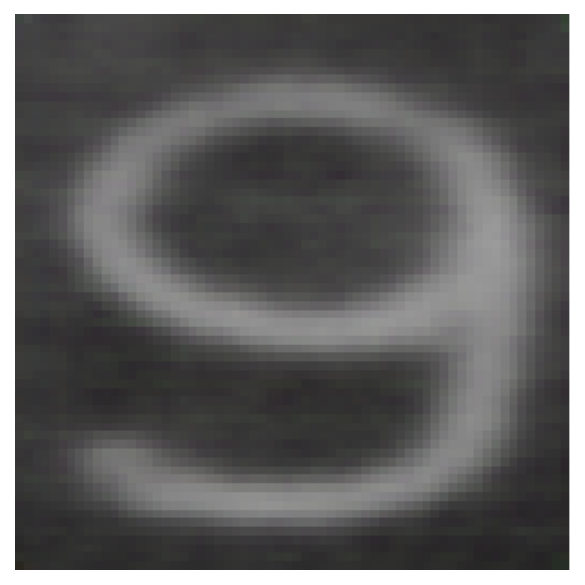}
        \caption{}
        \label{fig:reconstruct_mse_ssim_per}
    \end{subfigure}
    
    \caption{(a) Original resized image, (b) Original inverted image, (c)-(f) Reconstructed images from various models.}
    \label{fig:various_model_performance}
\end{figure}

The performance of the trained ResVAE model was evaluated on 500 individual character images, which were separated into two distinct sets. One set contained 250 non-defective characters, while the other included 250 defective characters exhibiting missing edges or incomplete prints. Figure \ref{fig:original_resized} provides an example of defective characters used for testing. These images were extracted from different nameplates and were not augmented, ensuring the evaluation closely reflects real-world conditions.  

\paragraph{Pre-processing}

Prior to being passed to the model, all images underwent pre-processing to ensure alignment and compatibility with the input requirements of the ResVAE model. The raw image samples varied in dimension, necessitating resizing and normalization. Each image was resized to \(64 \times 64\) pixels and rescaled to a range between 0.0 and 1.0. This resizing ensured compatibility with the ResVAE model and facilitated the use of the custom network for computing the feature perceptual loss. As described in the Feature Perceptual Loss section \ref{para:Feature Perceptual Loss}, this custom network employs convolutional layers initialized with pre-trained weights from VGG19, which can process input images of this size effectively. Additionally, pixel values were normalized to the range of \([-1.0, 1.0]\) to enhance the model's generalization ability.

All images were initially grayscale with a single channel. Since the images were captured using a camera with an optical filter mounted on the lens, they were inverted, as shown in Figure~\ref{fig:original_resized}. Before feeding them into the model, the images were re-inverted to restore their original appearance, as illustrated in Figure~\ref{fig:original_inverted}.

To match the input requirements of the neural network, especially given that the pre-trained weights of the VGG19 network were trained on the 3-channel ImageNet-1k dataset, each grayscale image was converted to an RGB format. This was achieved by replicating the single grayscale channel three times, resulting in a final input shape of \(64 \times 64 \times 3\). This transformation ensured compatibility with the pre-trained VGG19 weights, avoiding shape mismatches during training.

\paragraph{Training}

The ResVAE model was trained for 250 epochs using a batch size of 64 and a learning rate of 0.001. The training process was carried out iteratively, with the total loss function, defined in Equation \ref{eq:total_loss}, guiding the optimization of the model's performance. The PyTorch library was used to implement the model, and training was executed on an Azure platform equipped with four Nvidia T4 GPUs, ensuring efficient parallel processing and reduced training times.

The primary objective during training was to minimize the total loss for non-defective images, enabling the model to learn to accurately reconstruct normal characters. The training dataset consisted of both defective and non-defective images, while the ground truth dataset contained only non-defective images, serving as a clean reference for reconstruction. The total loss was computed based on the difference between the reconstructed image and its corresponding ground-truth image. As training progressed, the total loss gradually decreased, indicating the model’s improved ability to reconstruct non-defective characters.

\begin{figure}[htbp]
    \centering
    \captionsetup[subfigure]{justification=centering} 
    
    \begin{subfigure}{0.15\textwidth}
        \includegraphics[width=\linewidth]{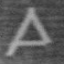}
        \caption{}
        \label{fig:Original image with defect}
    \end{subfigure} \hspace{1em}
    \begin{subfigure}{0.15\textwidth}
        \includegraphics[width=\linewidth]{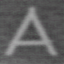}
        \caption{}
        \label{fig:reconstructed}
    \end{subfigure} \hspace{1em}
    \begin{subfigure}{0.15\textwidth}
        \includegraphics[width=\linewidth]{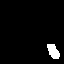}
        \caption{}
        \label{fig:anomaly_mask}
    \end{subfigure} \hspace{1em}
    \begin{subfigure}{0.15\textwidth}
        \includegraphics[width=\linewidth]{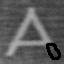}
        \caption{}
        \label{fig:contour_detection}
    \end{subfigure} 
    
    \caption{Anomaly Mask method steps. (a) Original image with defect, (b) Reconstructed image, (c) Anomaly mask, (d) Localized defects.}
    \label{fig:anomaly_steps}
\end{figure}

\paragraph{Anomaly Detection Methods}

In the previous section, the training process and the losses used to optimize the model were discussed, along with how the model reconstructs a non-defective character from a defective input. Building upon this, the next step involves localizing defects at the character level to identify anomalies. Therefore, two distinct approaches are employed for anomaly detection: the \textbf{Traditional Method} and the \textbf{Anomaly Mask Method}.

The \textbf{Traditional Method} involves classifying defective and non-defective images by calculating the reconstruction loss using Mean Squared Error (MSE). This loss is computed between the reconstructed image and the test image, as described in Equation \ref{eq:mse_loss}. A regularity score or threshold is applied, where images exceeding the threshold are classified as anomalous, while those with lower reconstruction loss are considered normal. Although straightforward, MSE-based classification primarily focuses on large reconstruction errors, often overlooking subtle defects \cite{khare2021analysis}. Additionally, this method can detect anomalies but lacks the capability to localize the defects on the character, limiting its effectiveness for precise defect identification.  

To overcome the limitations of the traditional method, an alternative \textbf{Anomaly Mask Method} was implemented. This method not only classifies defective characters but also localizes the defects by generating an anomaly mask, which is a binary image highlighting defective regions in white against a black background. The anomaly mask is created by performing a pixel-wise difference between the original image, as illustrated in Figure \ref{fig:Original image with defect}, and the reconstructed image, shown in Figure \ref{fig:reconstructed}. Pixels with error values above a defined threshold are set to white, while those below the threshold are set to black, resulting in a binary image representing defect regions, as depicted in Figure \ref{fig:anomaly_mask}.  

\begin{align}
A(i, j) = \begin{cases}
    1, & |I_{orig}(i, j) - I_{rec}(i, j)| > T \\
    0, & \text{otherwise}
\end{cases}
\label{eq:anomaly_mask}
\end{align}
where \( A(i, j) \) represents the anomaly mask, \( I_{orig} \) is the original image, \( I_{rec} \) is the reconstructed image, and \( T \) is the threshold.  

However, the anomaly mask generated through this method can introduce noise and artifacts, negatively impacting performance. To mitigate this, morphological operations \cite{Morphological} such as erosion and dilation are applied to refine the mask by removing small, unwanted regions and enhancing the visibility of the defect. Once the noise is eliminated and the defect is clearly highlighted, connected component analysis \cite{connectedopencv} is applied to the anomaly mask to isolate regions of interest. The size of each defect is then measured against a threshold, allowing for the classification of acceptable and unacceptable defects \cite{connectedcomp}.  

Defects that are below a threshold are classified as anomalies, and contours are drawn around these detected regions to localize and visualize the defects, as demonstrated in Figure \ref{fig:contour_detection}. The optimal thresholds for both the anomaly mask and connected component analysis were determined through a grid search process. Various combinations of thresholds were evaluated iteratively to identify the combination that maximized performance across key metrics, such as F1-score, while ensuring the recall metric achieved a value of 1.  

A recall value of 1 indicates that all defective characters are correctly identified, ensuring that no anomalies are missed. This is particularly critical in industrial applications, where even a single undetected defect can lead to significant quality control issues or operational failures. Prioritizing recall minimizes the risk of overlooking defective prints, aligning with the stringent requirements for defect detection in manufacturing environments. The F1-score, which represents the balance between precision and recall, is used to ensure that the selected thresholds not only detect all anomalies but also minimize the number of false positives, maintaining efficiency in the quality control pipeline.

\subsubsection{Experiments and Results}

\paragraph{Experiments}  

A series of experiments were conducted to evaluate the influence of each loss term on the quality of the reconstructed images. As illustrated in Figure \ref{fig:various_model_performance}, the experiments explore how individual and combined loss terms affect the reconstruction quality of the model.  

The image shown in Figure \ref{fig:reconstruct_mse} represents the result of training the model using only MSE and KL loss terms. This combination produced a blurred reconstruction, failing to accurately reconstruct the character. To improve structural integrity, an additional SSIM loss term was incorporated. The resulting reconstruction, shown in Figure \ref{fig:reconstruct_mse_ssim}, demonstrates enhanced structural information but a smoother background, indicating the model's focus on perceptual quality rather than sharpness.  

Further improvements were attempted by incorporating perceptual loss alongside MSE and KL loss terms. This configuration, depicted in Figure \ref{fig:reconstruct_mse_feature}, allowed the model to capture fine perceptual details; however, it failed to preserve the structural form of the character, leading to incomplete reconstructions.  

To leverage the benefits of each loss term, a combination of MSE, KL, SSIM, and perceptual loss was applied. The reconstructed image, as shown in Figure \ref{fig:reconstruct_mse_ssim_per}, closely resembles the input image, preserving both structural and perceptual information while reducing blurriness. However, as discussed in Section \ref{loss function}, using all loss terms without individual weighting can negatively impact training stability. To address this, weighted loss terms were applied, assigning individual weights to each component to balance their contributions.  

The loss weights were normalized to ensure they summed to 1, as expressed by the equation:  

\begin{equation}
\alpha + \beta + \gamma + \kappa = 1
\label{eq:loss_weight_sum}
\end{equation}
where \(\alpha\), \(\beta\), \(\gamma\), and \(\kappa\) represent the weights assigned to the MSE loss, KL divergence loss, SSIM loss, and perceptual loss, respectively.

This normalization simplifies hyperparameter tuning by providing a systematic approach to adjust the influence of each loss term. By redistributing the total weight among the loss components, the model can dynamically focus on critical aspects of reconstruction. Weights were predefined before training to ensure they consistently summed to 1 throughout the process.  

The selected weights, shown in Table \ref{tab:model_performance}, were determined through manual tuning. This process involved iteratively testing different weight combinations and selecting models that demonstrated better reconstruction quality and higher accuracy on performance metrics, such as recall, precision, and F1 score, on the validation set. In all experiments, the weight assigned to the MSE loss played a significant role, as it emphasizes pixel-wise differences and contributes heavily to the overall training objective.

In addition to these experiments, Models 5 and 6 incorporated a linear KL annealing strategy to mitigate posterior collapse in the latent space \cite{ichikawa2024learning}. Among these, Model 6 achieved the highest recall, benefiting from the linear KL annealing process. During KL annealing, the weight of the KL loss term was gradually increased while the weights of the other loss terms remained constant at their preset values. After reaching the maximum epoch for KL annealing, all loss weights were adjusted to satisfy the condition defined by Equation \ref{eq:loss_weight_sum}, ensuring that their sum equaled 1.

In Model 6, the KL loss weight started at 0.03 and progressively increased to 0.3 over 200 epochs. This gradual adjustment allowed the model to prioritize reconstruction quality during the initial stages of training. As the KL weight increased, the model shifted its focus toward learning a more regularized and meaningful latent space. This process enhanced the model's generalization capabilities and improved its stability during inference.

\begin{table}[h]
\centering
\caption{Performance Metrics and Hyperparameters for ResVAE-Based Anomaly Detection Using Traditional Method}
\label{tab:model_performance}
\renewcommand{\arraystretch}{1.3} 
\resizebox{\textwidth}{!}{%
\begin{tabular}{l c c c c c}
\hline
\textbf{Models} & \textbf{AUC (\%)} & \textbf{Precision (\%)} & \textbf{Recall (\%)} & \textbf{F1 Score (\%)} & \textbf{Accuracy (\%)} \\ 
\hline
\textbf{Model 1} ($\alpha=0.27, \beta=0.27, \gamma=0.22, \kappa=0.22$) & \textbf{99.3} & \textbf{91.1} & \textbf{98.8} & \textbf{94.8} & \textbf{94.6} \\ 
Model 2 ($\alpha=0.318, \beta=0.318, \gamma=0.136, \kappa=0.227$) & 97.6 & 90.3 & 96.8 & 93.4 & 93.2 \\ 
Model 3 ($\alpha=0.338, \beta=0.166, \gamma=0.1925, \kappa=0.2535$) & 98.7 & 91.0 & 97.6 & 94.2 & 94.0 \\ 
Model 4 ($\alpha=0.4, \beta=0.3, \gamma=0.0, \kappa=0.3$) & 97.0 & 90.5 & 95.2 & 92.8 & 92.6 \\ 
Model 5 ($\alpha=0.35, \beta=0.07\text{-}0.3, \gamma=0.15, \kappa=0.2$) & 99.1 & 90.7 & 97.2 & 93.8 & 93.6 \\ 
Model 6 ($\alpha=0.4, \beta=0.03\text{-}0.3, \gamma=0.2, \kappa=0.1$) & 99.4 & 83.9 & 100.0 & 91.2 & 90.4 \\ 
\hline
\end{tabular}%
}
\end{table}

\paragraph{Results}  

To evaluate the performance of the models, five independent metrics were computed. The primary metric used was the Area Under the Receiver Operating Characteristic curve (AUC), which measures the model's ability to distinguish between defective and non-defective images. AUC reflects the True Positive Rate (TPR), representing the percentage of correctly classified defective images, and the False Positive Rate (FPR), which is the proportion of non-defective images incorrectly classified as defective. A higher AUC value indicates better discrimination between defective and non-defective images. In addition to AUC, other key metrics included accuracy, precision, recall, and F1 score. 

The models, as detailed in Table \ref{tab:model_performance}, were evaluated using these metrics to identify the best-performing configuration. For the traditional method, thresholds were determined by computing the mean squared error (MSE) between the reconstructed and input images, as described in Equation \ref{eq:mse_loss}. The threshold serves as the decision boundary, classifying images with low reconstruction error as non-defective and those with high reconstruction error as defective.  

The optimal threshold was selected by analyzing the Receiver Operating Characteristic (ROC) curve. The threshold was chosen based on specific requirements, prioritizing a high true positive rate while minimizing the false positive rate. Specifically, the point where the true positive rate approaches 1 and the false positive rate is minimized, ideally close to 0, was selected. This requirement ensures that recall is prioritized, reducing the likelihood of defective items being overlooked, which is critical in industrial applications where undetected defects can compromise quality and safety.  

While this method effectively identifies positive cases, the emphasis on achieving a high recall often results in an increased number of false positives, leading to the misclassification of non-defective items as defective. This trade-off between recall and precision is evident across the models, with Model 1 demonstrating superior performance across all metrics. As shown in Table \ref{tab:model_performance}, Model 1 achieves the highest accuracy and provides a balanced trade-off between precision and recall, making it the most reliable configuration for defect detection.  

Despite these strengths, achieving perfect recall and precision simultaneously using the traditional method remains challenging when relying solely on MSE as the reconstruction error. MSE calculates the overall pixel-level difference between the input and reconstructed images by averaging these differences across the entire image. In cases where most regions are accurately reconstructed, small or localized defects may contribute only a minor increase in the overall error. This averaging effect can effectively hide subtle anomalies, meaning that even a clearly defective area might not push the total error above the detection threshold. Lowering the threshold to capture these small defects, however, risks increasing the false positive rate by classifying normal variations, such as minor texture differences or slight lighting inconsistencies, as defects. This inherent trade-off makes it difficult for a single, global MSE-based threshold to capture all true anomalies while simultaneously avoiding the misclassification of normal items. 

To address the limitations of the traditional method, the \textbf{Anomaly Mask Method} was implemented, significantly enhancing both anomaly detection and defect localization. Model 1, which demonstrated balanced performance across all metrics, was used as the baseline for applying the anomaly mask method. This approach led to improved results, achieving an accuracy of 99.8\%, recall of 100\%, precision of 99.6\%, and an F1 score of 99.8\%.   

\paragraph{Discussion}  

The anomaly mask method showed higher accuracy and better overall performance compared to the traditional approach. However, a minor limitation was observed, as one non-defective image was misclassified as defective, resulting in a slight reduction in the precision score. This misclassification highlights the sensitivity of the model to variations in brightness between the original and reconstructed images. Specifically, when the brightness of the captured characters is higher than that of the reconstructed images, the pixel-wise differencing process can introduce additional noise. This noise amplifies minor discrepancies, increasing the likelihood of false positives.  

To mitigate the impact of varying lighting conditions and enhance the model's robustness, techniques such as histogram equalization can be employed. By normalizing the brightness and contrast of both the original and reconstructed images, histogram equalization reduces inconsistencies caused by uneven lighting. This pre-processing step can effectively minimize noise during the differencing process, leading to more accurate anomaly detection and reducing false positive rates. Implementing this enhancement has the potential to significantly improve the model’s reliability across diverse operational environments, ensuring consistent performance even under fluctuating lighting conditions.  
 
\subsection{Working of the Whole Pipeline}  

The primary objective of the proposed pipeline is to detect and localize anomalies on nameplates. To achieve this, the models and methods discussed in previous sections are integrated into a comprehensive system designed to fulfill the requirements for defect detection and anomaly localization on characters and logos. Figure \ref{fig:pipeline} illustrates the complete pipeline, which consolidates all models into a unified process.  

\subsubsection{Working Principle}

Each component within the pipeline plays a distinct role in ensuring the accuracy and correctness of the laser-engraved data on a nameplate. The overall decision-making process of the pipeline is summarized in Algorithm~\ref{alg:nameplate_inspection}, which outlines how a nameplate is classified as either \textit{acceptable} or \textit{defective} based on sequential verification steps.

The process begins with image alignment, which requires a reference image of the same type of nameplate. When a new image is captured by the camera during production, the initial step is to align the reference image with the newly captured one, as described in Section~\ref{subsec: Image Alignment and Logo Defect Detection}. This alignment process ensures that the two images are geometrically synchronized, minimizing positional discrepancies that could interfere with downstream tasks.

Following alignment, the YOLO model processes both the aligned reference image and the captured image to detect and extract strings, logos, and the Data Matrix Code (DMC). The model generates two sets of bounding box coordinates, one for the aligned reference image and the other for the captured image. These bounding boxes from the aligned reference image are used to locate and extract corresponding logos from the captured image, ensuring that both images are analyzed in the same regions. The extracted logos from the aligned reference and captured images are then compared to identify discrepancies or defects, as detailed in Section~\ref{subsec: Image Alignment and Logo Defect Detection}. If any inconsistencies are detected at this stage, the nameplate is classified as \textit{defective}, and further analysis is halted.

If no anomalies are found in the logos, the pipeline proceeds to the next stage, where the strings extracted from the captured image are processed using Optical Character Recognition (OCR). This step, explained in Section~\ref{subsec: string detection and recognition}, facilitates the detection and recognition of individual characters within the strings. Once the OCR model outputs the recognized strings, they are compared against the expected values stored in the MES. If any discrepancies are found in this comparison, the nameplate is immediately classified as \textit{defective}.

Following string verification, individual characters are passed to the autoencoder for further anomaly detection at the character level. As outlined in Section~\ref{subsec: anomaly detection on single characters}, the autoencoder identifies subtle defects that may not have been captured in previous stages. Each character is reconstructed by the autoencoder, and post-processing is applied to detect and localize any anomalies. If the autoencoder detects defects, the nameplate is classified as \textit{defective}, and further inspection steps are halted.

The pipeline systematically examines the entire nameplate, beginning with logos and progressing to character-level verification. If all verification stages—logo defect detection, string recognition, and character-level anomaly detection—are completed without identifying any defects, the nameplate is classified as \textit{acceptable}. However, if defects are detected at any stage, the nameplate is categorized as \textit{defective} and does not meet quality standards.
  
This hierarchical and multi-stage pipeline ensures that no defective nameplates proceed further in the production process, maintaining high quality control standards and minimizing the risk of defective parts being dispatched to customers.

\subsubsection{Experiments and Results}  

\paragraph{Experimental Setup}  

The performance evaluation of the entire pipeline was performed on a dataset comprising 150 nameplates. This dataset was evenly divided into 75 non-defective nameplates and 75 defective nameplates. Each nameplate image had a resolution of 1920$\times$1600, containing both strings and logos. The defective nameplate dataset included samples exhibit various types of anomalies, such as character defects, logo defects, and missing strings. To further assess the robustness of the pipeline, the evaluation dataset also contained images captured under different lighting conditions, simulating real-world variability in production environments. This ensured that the pipeline was tested not only for defect detection, but also for its resilience to changes in image quality caused by external factors.  

\paragraph{Results}  

The performance of the entire pipeline was evaluated using common metrics, including accuracy, precision, recall, and F1 score. A nameplate was classified as defective if any of the integrated models identified anomalies. The models contributing to the pipeline's decision included YOLO for string and logo detection, the logo defect detection method, Tesseract OCR for string verification, and the residual variational autoencoder for character-level anomaly detection.  

Upon evaluation, the pipeline achieved an overall accuracy of 91.33\%, with a precision of 85.23\%, a recall of 100\%, and an F1 score of 92.02\%. The pipeline successfully detected all defective nameplates, resulting in no false negatives. However, 13 non-defective nameplates were misclassified as defective, contributing to the reduced precision and overall accuracy. Despite these misclassifications, the pipeline demonstrated high recall, ensuring that no defective nameplates went undetected. 

\paragraph{Discussion}  

The results indicate that the proposed pipeline effectively detects and localizes defects on nameplates. However, the limitations of individual models, as addressed in the respective discussion sections, contributed to the misclassification of certain non-defective nameplates. This trade-off between precision and recall highlights the challenges faced by the pipeline in maintaining perfect classification accuracy across all conditions.  

Among the 13 misclassified non-defective nameplates, the Tesseract OCR model incorrectly classified 3 samples due to the insertion of extra characters, leading to string mismatches during comparison. The residual variational autoencoder misclassified 6 nameplates, primarily due to darker images. Under these conditions, the post-processing method applied to the anomaly mask introduced noise, resulting in false positives. Additionally, the logo defect detection module misclassified 4 nameplates, as surface anomalies and variations in brightness negatively impacted the performance of the traditional computer vision techniques used for defect detection. Notably, YOLO exhibited no misclassifications, successfully detecting all logos and strings with high accuracy. While the pipeline’s high recall ensures that defective nameplates are consistently identified, the reduction in precision highlights the need to minimize false positives. 
 
\begin{table}[h]
    \centering
    \caption{Performance Metrics for Different Stages of the Pipeline. WLA: Word-Level Accuracy, CLA: Character-Level Accuracy.}
    \label{tab:performance_metrics_single}
    \renewcommand{\arraystretch}{1.2} 
    \begin{tabular}{l c c c c}
        \hline
        \textbf{Name}                              & \textbf{Accuracy (\%)}         & \textbf{Precision (\%)} & \textbf{Recall (\%)} & \textbf{F1 Score (\%)} \\
        \hline
        \textbf{String Detection}                  & 100.0                         & 100.0                   & 100.0                & 100.0                  \\
        \textbf{Logo Defect Detection}             & 97.9                          & 95.5                    & 100.0                & 97.7                   \\
        \textbf{Character Detection/Recognition}   & \textbf{WLA:} 98.7  \textbf{CLA:} 99.7  & N/A                     & N/A                  & N/A                    \\
        \textbf{Anomaly Detection on Characters}   & 99.8                          & 99.6                    & 100.0                & 99.8                   \\
        \hline
    \end{tabular}
\end{table}

\section{Conclusion}  

In conclusion, this research introduces a comprehensive pipeline designed to detect and localize defects on laser-engraved nameplates by integrating multiple models and computer vision techniques. The system combines the YOLOv7 model for string and logo detection and common computer vision techniques for logo inspection, Tesseract OCR for character recognition and detection, and a residual variational autoencoder for character-level anomaly detection, addressing a wide range of potential defects including missing or misprinted characters and logo inconsistencies. Figure \ref{fig:image2} illustrates the final output of the pipeline, showcasing its capability to identify and localize anomalies across the nameplate. This integrated approach enhances production quality by providing an automated solution to nameplate inspection, reducing the reliance on manual quality control.  

The performance evaluation of the full pipeline demonstrates its effectiveness in defect detection, achieving an overall accuracy of 91.33\%, with a precision of 85.23\%, a recall of 100\%, and an F1 score of 92.02\%. Table \ref{tab:performance_metrics_single} highlights the contributions of each stage of the pipeline, with the anomaly detection and logo defect modules achieving 100\% recall. This ensures that all defective nameplates were correctly identified, reflecting the system’s reliability in capturing and localizing defects at different stages of the process. The high recall rate reinforces the pipeline’s ability to detect even subtle anomalies, minimizing the risk of defective nameplates passing through inspection.  

Despite the strong recall performance, the pipeline demonstrated limitations in precision, resulting in 13 non-defective nameplates being misclassified as defective. This misclassification was mainly due to the residual variational autoencoder sensitivity to darker images and the dependence of the logo defect detection module on traditional computer vision techniques, which were affected by surface irregularities and variations in brightness. Addressing these misclassifications is crucial for improving overall precision without compromising recall.  

Future work will focus on enhancing the robustness of the autoencoder and the logo defect detection module under varying lighting and surface conditions. To mitigate the impact of brightness variations and surface artifacts, strategies such as adaptive thresholding, improved post-processing techniques, and the integration of machine learning-based anomaly detection methods will be explored.

Additionally, the pre-processing step for Tesseract OCR in this study relied on traditional binarization techniques, which can introduce noise and negatively affect performance. A potential improvement involves replacing this approach with a convolutional neural network-based binarization method, such as DP-LinkNet, proposed by Xiong \textit{et al.} \cite{xiong2021dp}. DP-LinkNet combines hybrid dilated convolution and spatial pyramid pooling to enhance text binarization, which could lead to improved OCR accuracy.

Furthermore, the dependency on Tesseract OCR for text recognition in the pipeline could be reconsidered. An alternative approach involves replacing Tesseract with CRAFT (Character Region Awareness for Text Detection) \cite{baek2019character}. However, since CRAFT focuses solely on text detection, an additional text recognition model would need to be integrated. Integrating additional neural networks or models in the pipeline could increase inference time, which must be carefully evaluated to maintain efficiency in production settings.

Lastly, the current pipeline does not incorporate DMC reading, which could be seamlessly integrated using a dedicated DMC reader library \cite{DMC}. Implementing these improvements will enhance the overall accuracy and robustness of the system, reducing false positives and increasing reliability across different production environments while ensuring that any additional computational overhead remains manageable.  

The proposed pipeline represents a significant step toward automating nameplate inspection and defect detection, particularly in controlled industrial environments. With continued development and optimization, the system has the potential to streamline quality control processes, reduce inspection times, and improve overall production efficiency.

\section*{Acknowledgment}
\textit{The authors wish to express sincere gratitude to Knorr-Bremse SfN GmbH at plant Aldersbach for providing the opportunity, resources, and collaborative environment necessary for this research. Special thanks go to team lead Gerhard Maier, unit manager Andreas Panitschka, and plant manager Gerhard Schwarz for their insightful feedback, leadership, and encouragement during the research process.}

\textit{Furthermore, the author is deeply grateful to Prof. Noah Klarmann at Technische Hochschule Rosenheim for their academic supervision, mentorship, and continuous support. This collaboration between academia and industry has been a pivotal factor in the successful completion of this research and the resulting publication.}

\bibliography{references}

\begin{thebibliography}{10}

\bibitem{wang2023yolov7}
Chien-Yao Wang, Alexey Bochkovskiy, and Hong-Yuan~Mark Liao.
\newblock Yolov7: Trainable bag-of-freebies sets new state-of-the-art for real-time object detectors.
\newblock In {\em Proceedings of the IEEE/CVF conference on computer vision and pattern recognition}, pages 7464--7475, 2023.

\bibitem{kim2018smart}
Dong-Hyeon Kim, Thomas~JY Kim, Xinlin Wang, Mincheol Kim, Ying-Jun Quan, Jin~Woo Oh, Soo-Hong Min, Hyungjung Kim, Binayak Bhandari, Insoon Yang, et~al.
\newblock Smart machining process using machine learning: A review and perspective on machining industry.
\newblock {\em International Journal of Precision Engineering and Manufacturing-Green Technology}, 5:555--568, 2018.

\bibitem{liao2019state}
Zhirong Liao, Ali Abdelhafeez, Haonan Li, Yue Yang, Oriol~Gavalda Diaz, and Dragos Axinte.
\newblock State-of-the-art of surface integrity in machining of metal matrix composites.
\newblock {\em International journal of machine tools and manufacture}, 143:63--91, 2019.

\bibitem{xie2008review}
Xianghua Xie.
\newblock A review of recent advances in surface defect detection using texture analysis techniques.
\newblock {\em ELCVIA: electronic letters on computer vision and image analysis}, pages 1--22, 2008.

\bibitem{aslam2020comparative}
Yasir Aslam and N~Santhi.
\newblock A comparative study of thresholding based defect detection techniques.
\newblock In {\em Intelligent Data Communication Technologies and Internet of Things: ICICI 2019}, pages 631--637. Springer, 2020.

\bibitem{kumar2008computer}
Ajay Kumar.
\newblock Computer-vision-based fabric defect detection: A survey.
\newblock {\em IEEE transactions on industrial electronics}, 55(1):348--363, 2008.

\bibitem{8278352}
Binwu Ma, Wei Zhu, Yanghong Wang, Huan Wu, Yanzhu Yang, Hui Fan, and Hongwei Xu.
\newblock The defect detection of personalized print based on template matching.
\newblock In {\em 2017 IEEE International Conference on Unmanned Systems (ICUS)}, pages 266--271, 2017.

\bibitem{ce2017pcb}
Win Ce et~al.
\newblock Pcb defect detection using opencv with image subtraction method.
\newblock In {\em 2017 International Conference on Information Management and Technology (ICIMTech)}, pages 204--209. IEEE, 2017.

\bibitem{yang2020using}
Jing Yang, Shaobo Li, Zheng Wang, Hao Dong, Jun Wang, and Shihao Tang.
\newblock Using deep learning to detect defects in manufacturing: a comprehensive survey and current challenges.
\newblock {\em Materials}, 13(24):5755, 2020.

\bibitem{9873703}
Jing Li, Xiaoli Bai, Jie Pan, Quanhui Tian, Wanying Fu, and Zhaohui Jing.
\newblock A deep learning method for printing defect detection.
\newblock In {\em 2022 IEEE 4th International Conference on Power, Intelligent Computing and Systems (ICPICS)}, pages 246--249, 2022.

\bibitem{7485869}
Shaoqing Ren, Kaiming He, Ross Girshick, and Jian Sun.
\newblock Faster r-cnn: Towards real-time object detection with region proposal networks.
\newblock {\em IEEE Transactions on Pattern Analysis and Machine Intelligence}, 39(6):1137--1149, 2017.

\bibitem{Detectron2}
Y. wu, a. kirillov, f. massa, w.-y. lo, and r. girshick, “detectron2,”.
\newblock \url{https://github.com/facebookresearch/detectron2}.
\newblock Accessed: 2024-10-06.

\bibitem{wang2023comprehensive}
Xiangheng Wang, Hengyi Li, Xuebin Yue, and Lin Meng.
\newblock A comprehensive survey on object detection yolo.
\newblock {\em Proceedings http://ceur-ws. org ISSN}, 1613:0073, 2023.

\bibitem{sharma2024optimizing}
Pravek Sharma, Rajesh Tyagi, and Priyanka Dubey.
\newblock Optimizing real-time object detection-a comparison of yolo models.
\newblock {\em International Journal of Innovative Research in Computer Science \& Technology}, 12(3):57--74, 2024.

\bibitem{raj2022comprehensive}
Ravi Raj and Andrzej Kos.
\newblock A comprehensive study of optical character recognition.
\newblock In {\em 2022 29th International Conference on Mixed Design of Integrated Circuits and System (MIXDES)}, pages 151--154. IEEE, 2022.

\bibitem{smith2007overview}
Ray Smith.
\newblock An overview of the tesseract ocr engine.
\newblock In {\em Ninth international conference on document analysis and recognition (ICDAR 2007)}, volume~2, pages 629--633. IEEE, 2007.

\bibitem{anwar2022text}
Nadeem Anwar, Tauseef Khan, and Ayatullah~Faruk Mollah.
\newblock Text detection from scene and born images: How good is tesseract?
\newblock In {\em Recent Trends in Communication and Intelligent Systems: Proceedings of ICRTCIS 2021}, pages 115--122. Springer, 2022.

\bibitem{brisinello2017improving}
Matteo Brisinello, Ratko Grbi{\'c}, Matija Pul, and Tihomir An{\dj}eli{\'c}.
\newblock Improving optical character recognition performance for low quality images.
\newblock In {\em 2017 International Symposium ELMAR}, pages 167--171. IEEE, 2017.

\bibitem{li2023trocr}
Minghao Li, Tengchao Lv, Jingye Chen, Lei Cui, Yijuan Lu, Dinei Florencio, Cha Zhang, Zhoujun Li, and Furu Wei.
\newblock Trocr: Transformer-based optical character recognition with pre-trained models.
\newblock In {\em Proceedings of the AAAI Conference on Artificial Intelligence}, volume~37, pages 13094--13102, 2023.

\bibitem{hu2017wordsup}
Han Hu, Chengquan Zhang, Yuxuan Luo, Yuzhuo Wang, Junyu Han, and Errui Ding.
\newblock Wordsup: Exploiting word annotations for character based text detection.
\newblock In {\em Proceedings of the IEEE international conference on computer vision}, pages 4940--4949, 2017.

\bibitem{baek2019character}
Youngmin Baek, Bado Lee, Dongyoon Han, Sangdoo Yun, and Hwalsuk Lee.
\newblock Character region awareness for text detection.
\newblock In {\em Proceedings of the IEEE/CVF conference on computer vision and pattern recognition}, pages 9365--9374, 2019.

\bibitem{tao2022deep}
Xian Tao, Xinyi Gong, Xin Zhang, Shaohua Yan, and Chandranath Adak.
\newblock Deep learning for unsupervised anomaly localization in industrial images: A survey.
\newblock {\em IEEE Transactions on Instrumentation and Measurement}, 71:1--21, 2022.

\bibitem{goodfellow2014generative}
Ian Goodfellow, Jean Pouget-Abadie, Mehdi Mirza, Bing Xu, David Warde-Farley, Sherjil Ozair, Aaron Courville, and Yoshua Bengio.
\newblock Generative adversarial nets.
\newblock {\em Advances in neural information processing systems}, 27, 2014.

\bibitem{chow2020anomaly}
Jun~Kang Chow, Zhaoyu Su, Jimmy Wu, Pin~Siang Tan, Xin Mao, and Yu-Hsing Wang.
\newblock Anomaly detection of defects on concrete structures with the convolutional autoencoder.
\newblock {\em Advanced Engineering Informatics}, 45:101105, 2020.

\bibitem{shi2023lightweight}
Hui Shi, Gangyan Li, and Hanwei Bao.
\newblock Lightweight reconstruction network for surface defect detection based on texture complexity analysis.
\newblock {\em Electronics}, 12(17):3617, 2023.

\bibitem{kingma2013auto}
Diederik~P Kingma.
\newblock Auto-encoding variational bayes.
\newblock {\em arXiv preprint arXiv:1312.6114}, 2013.

\bibitem{chen2023auto}
Shuangshuang Chen and Wei Guo.
\newblock Auto-encoders in deep learning—a review with new perspectives.
\newblock {\em Mathematics}, 11(8):1777, 2023.

\bibitem{he2022survey}
Xiangjie He, Zhengwei Chang, Linghao Zhang, Houdong Xu, Hongbo Chen, and Zhongqiang Luo.
\newblock A survey of defect detection applications based on generative adversarial networks.
\newblock {\em IEEE Access}, 10:113493--113512, 2022.

\bibitem{xianxu2024feature}
Guoping~Qiu Xianxu~Hou, Linlin~Shen.
\newblock Feature perceptual loss for variational autoencoder.
\newblock {\em arXiv preprint arXiv:1610.00291}, 2024.

\bibitem{10104785}
Ankit Kumar and Manju Khari.
\newblock Efficient video anomaly detection using residual variational autoencoder.
\newblock In {\em 2023 International Conference on Communication System, Computing and IT Applications (CSCITA)}, pages 50--55, 2023.

\bibitem{you2022unified}
Zhiyuan You, Lei Cui, Yujun Shen, Kai Yang, Xin Lu, Yu~Zheng, and Xinyi Le.
\newblock A unified model for multi-class anomaly detection.
\newblock {\em Advances in Neural Information Processing Systems}, 35:4571--4584, 2022.

\bibitem{he2024diffusion}
Haoyang He, Jiangning Zhang, Hongxu Chen, Xuhai Chen, Zhishan Li, Xu~Chen, Yabiao Wang, Chengjie Wang, and Lei Xie.
\newblock A diffusion-based framework for multi-class anomaly detection.
\newblock In {\em Proceedings of the AAAI Conference on Artificial Intelligence}, volume~38, pages 8472--8480, 2024.

\bibitem{6999108}
Jingmei Li, Weiguo Zhang, and Ruili Han.
\newblock Application of machine vision in defects inspection and character recognition of nameplate surface.
\newblock In {\em 2014 13th International Symposium on Distributed Computing and Applications to Business, Engineering and Science}, pages 295--298, 2014.

\bibitem{peng2021defect}
Jianzhong Peng, Wei Zhu, Qiaokang Liang, Zhengwei Li, Maoying Lu, Wei Sun, and Yaonan Wang.
\newblock Defect detection in code characters with complex backgrounds based on bbe.
\newblock {\em Mathematical Biosciences and Engineering}, 18(4):3755--3780, 2021.

\bibitem{liu2023printing}
Xinyu Liu, Yao Li, Yiyu Guo, and Luoyu Zhou.
\newblock Printing defect detection based on scale-adaptive template matching and image alignment.
\newblock {\em Sensors}, 23(9):4414, 2023.

\bibitem{elanangai2019automated}
V~Elanangai and Vasanth~Kishore Babu.
\newblock Automated system for defect identification and character recognition using ir images of ss-plates.
\newblock {\em International Journal of Recent Technology and Engineering}, 8(3), 2019.

\bibitem{haitao2023surface}
X~Haitao, P~Haipeng, and L~Junfeng.
\newblock Surface defect detection of bearing rings based on an improved yolov5 network.[j].
\newblock {\em Sensors (Basel, Switzerland)}, 23(17), 2023.

\bibitem{cutts1997introduction}
Matt Cutts.
\newblock An introduction to the gimp.
\newblock {\em XRDS: Crossroads, The ACM Magazine for Students}, 3(4):28--30, 1997.

\bibitem{karami2017image}
Ebrahim Karami, Siva Prasad, and Mohamed Shehata.
\newblock Image matching using sift, surf, brief and orb: performance comparison for distorted images.
\newblock {\em arXiv preprint arXiv:1710.02726}, 2017.

\bibitem{7827292}
Frazer~K. Noble.
\newblock Comparison of opencv's feature detectors and feature matchers.
\newblock In {\em 2016 23rd International Conference on Mechatronics and Machine Vision in Practice (M2VIP)}, pages 1--6, 2016.

\bibitem{BFMatcher}
Opencv brute-force descriptor matcher documentation.
\newblock \url{https://docs.opencv.org/3.4/d3/da1/classcv_1_1BFMatcher.html#details}.
\newblock Accessed: 2024-10-13.

\bibitem{zhang2008automatic}
Rong Zhang.
\newblock Automatic computation of a homography by ransac algorithm.
\newblock {\em ECE661 Computer Vision}, 2008.

\bibitem{bazargani2018fast}
Hamid Bazargani, Olexa Bilaniuk, and Robert Laganiere.
\newblock A fast and robust homography scheme for real-time planar target detection.
\newblock {\em Journal of Real-Time Image Processing}, 15(4):739--758, 2018.

\bibitem{Morphological}
Opencv morphological transformations documentation.
\newblock \url{https://docs.opencv.org/4.x/d9/d61/tutorial_py_morphological_ops.html}.
\newblock Accessed: 2024-10-06.

\bibitem{connectedopencv}
Opencv connected component analysis documentation.
\newblock \url{https://docs.opencv.org/4.x/d0/d05/group__cudaimgproc.html#ga65b804fa23149c0fc54c5586dab04c68}.
\newblock Accessed: 2024-10-07.

\bibitem{connectedcomp}
Connected component analysis documentation.
\newblock \url{https://pyimagesearch.com/2021/02/22/opencv-connected-component-labeling-and-analysis/}.
\newblock Accessed: 2024-10-07.

\bibitem{gittess}
Tesseract official github.
\newblock \url{https://github.com/tesseract-ocr/tesseract.git}.
\newblock Accessed: 2024-10-07.

\bibitem{tessgit}
Tesseract documentation.
\newblock \url{https://tesseract-ocr.github.io/tessdoc/ImproveQuality}.
\newblock Accessed: 2024-10-08.

\bibitem{yolov7_hyp}
Yolo official github hyperparameter.
\newblock \url{https://github.com/WongKinYiu/yolov7/blob/main/data/hyp.scratch.custom.yaml}.
\newblock Accessed: 2024-10-08.

\bibitem{tessmodel}
Tesseract trained data file.
\newblock \url{https://tesseract-ocr.github.io/tessdoc/Data-Files}.
\newblock Accessed: 2024-10-07.

\bibitem{nltk}
Natural language toolkit documentation.
\newblock \url{https://www.nltk.org/api/nltk.metrics.edit_distance.html}.
\newblock Accessed: 2024-10-07.

\bibitem{kingma2019introduction}
Diederik~P Kingma, Max Welling, et~al.
\newblock An introduction to variational autoencoders.
\newblock {\em Foundations and Trends{\textregistered} in Machine Learning}, 12(4):307--392, 2019.

\bibitem{borawar2023resnet}
Lokesh Borawar and Ravinder Kaur.
\newblock Resnet: Solving vanishing gradient in deep networks.
\newblock In {\em Proceedings of International Conference on Recent Trends in Computing: ICRTC 2022}, pages 235--247. Springer, 2023.

\bibitem{appati2023deep}
Justice~Kwame Appati, Pius Gyamenah, Ebenezer Owusu, and Winfred Yaokumah.
\newblock Deep residual variational autoencoder for image super-resolution.
\newblock In {\em International Conference on Information, Communication and Computing Technology}, pages 91--103. Springer, 2023.

\bibitem{6026515}
Karl Leboeuf, Roberto Muscedere, and Majid Ahmadi.
\newblock Performance analysis of table-based approximations of the hyperbolic tangent activation function.
\newblock In {\em 2011 IEEE 54th International Midwest Symposium on Circuits and Systems (MWSCAS)}, pages 1--4, 2011.

\bibitem{9345813}
Amir~Hossein Estiri, Mohammad~Reza Sabramooz, Ali Banaei, Amir~Hossein Dehghan, Benyamin Jamialahmadi, and Mahdi~Jafari Siavoshani.
\newblock A variational auto-encoder approach for image transmission in noisy channel.
\newblock In {\em 2020 10th International Symposium onTelecommunications (IST)}, pages 227--233, 2020.

\bibitem{1284395}
Zhou Wang, A.C. Bovik, H.R. Sheikh, and E.P. Simoncelli.
\newblock Image quality assessment: from error visibility to structural similarity.
\newblock {\em IEEE Transactions on Image Processing}, 13(4):600--612, 2004.

\bibitem{nilsson2020understanding}
Jim Nilsson and Tomas Akenine-M{\"o}ller.
\newblock Understanding ssim.
\newblock {\em arXiv preprint arXiv:2006.13846}, 2020.

\bibitem{zhao2016loss}
Hang Zhao, Orazio Gallo, Iuri Frosio, and Jan Kautz.
\newblock Loss functions for image restoration with neural networks.
\newblock {\em IEEE Transactions on computational imaging}, 3(1):47--57, 2016.

\bibitem{johnson2016perceptual}
Justin Johnson, Alexandre Alahi, and Li~Fei-Fei.
\newblock Perceptual losses for real-time style transfer and super-resolution.
\newblock In {\em Computer Vision--ECCV 2016: 14th European Conference, Amsterdam, The Netherlands, October 11-14, 2016, Proceedings, Part II 14}, pages 694--711. Springer, 2016.

\bibitem{7926714}
Xianxu Hou, Linlin Shen, Ke~Sun, and Guoping Qiu.
\newblock Deep feature consistent variational autoencoder.
\newblock In {\em 2017 IEEE Winter Conference on Applications of Computer Vision (WACV)}, pages 1133--1141, 2017.

\bibitem{vgg19}
Vgg19 pytorch documentation.
\newblock \url{https://pytorch.org/vision/master/models/generated/torchvision.models.vgg19.html}.
\newblock Accessed: 2024-10-13.

\bibitem{simonyan2014very}
Karen Simonyan and Andrew Zisserman.
\newblock Very deep convolutional networks for large-scale image recognition.
\newblock {\em arXiv preprint arXiv:1409.1556}, 2014.

\bibitem{khare2021analysis}
Nishant Khare, Poornima~Singh Thakur, Pritee Khanna, and Aparajita Ojha.
\newblock Analysis of loss functions for image reconstruction using convolutional autoencoder.
\newblock In {\em International Conference on Computer Vision and Image Processing}, pages 338--349. Springer, 2021.

\bibitem{ichikawa2024learning}
Yuma Ichikawa and Koji Hukushima.
\newblock Learning dynamics in linear vae: Posterior collapse threshold, superfluous latent space pitfalls, and speedup with kl annealing.
\newblock In {\em International Conference on Artificial Intelligence and Statistics}, pages 1936--1944. PMLR, 2024.

\bibitem{xiong2021dp}
Wei Xiong, Xiuhong Jia, Dichun Yang, Meihui Ai, Lirong Li, and Song Wang.
\newblock Dp-linknet: A convolutional network for historical document image binarization.
\newblock {\em KSII Transactions on Internet and Information Systems (TIIS)}, 15(5):1778--1797, 2021.

\bibitem{DMC}
Read and write data matrix barcodes.
\newblock \url{https://pypi.org/project/pylibdmtx/}.
\newblock Accessed: 2024-10-08.

\end{thebibliography}

\clearpage
\pagestyle{empty} 
\renewcommand{\thispagestyle}[1]{} 
\begin{appendices}
\section{Appendix} 

\label{sec:appendix}

\subsection{Structured Decision Process for Nameplate Inspection}

\renewcommand{\algorithmicrequire}{\textbf{Input:}}
\renewcommand{\algorithmicensure}{\textbf{Output:}}

\begin{algorithm}[H]
\caption{Decision Flow for Nameplate Inspection}
\label{alg:nameplate_inspection}
\begin{algorithmic}[1]
\REQUIRE 
    - Captured image \( I_{\text{captured}} \) \\
    - Reference image \( I_{\text{reference}} \) \\
    - Expected string values from MES
\ENSURE 
    - Classification result: Acceptable or Defective

\STATE Align \( I_{\text{reference}} \) with \( I_{\text{captured}} \)
\STATE Use YOLO to detect logos, strings, and DMC in \( I_{\text{captured}} \) and \( I_{\text{reference}} \)
\STATE Extract logos from \( I_{\text{captured}} \) using bounding boxes from \( I_{\text{reference}} \)
\STATE Compare extracted captured image logos with reference logos
\IF{discrepancies in logos are found}
    \STATE Classify nameplate as Defective
    \STATE Exit
\ENDIF
\STATE Use Tesseract OCR to recognize strings in \( I_{\text{captured}} \)
\STATE Compare recognized strings with expected values from MES
\IF{discrepancies in strings are found}
    \STATE Classify nameplate as Defective
    \STATE Exit
\ENDIF
\FOR{each recognized character}
    \STATE Pass character to autoencoder for anomaly detection
    \IF{anomalies are found}
        \STATE Classify nameplate as Defective
        \STATE Exit
    \ENDIF
\ENDFOR
\STATE Classify nameplate as Acceptable

\end{algorithmic}
\end{algorithm}

\subsection{Supplementary Material}

This section provides supplementary material to support the data structure and methodology discussed in the paper. The figures below illustrate the diversity and complexity of the dataset used in the development of the proposed pipeline.

\begin{figure}[H]
    \centering
    \begin{subfigure}[t]{0.45\textwidth}
        \centering
        \includegraphics[width=\textwidth]{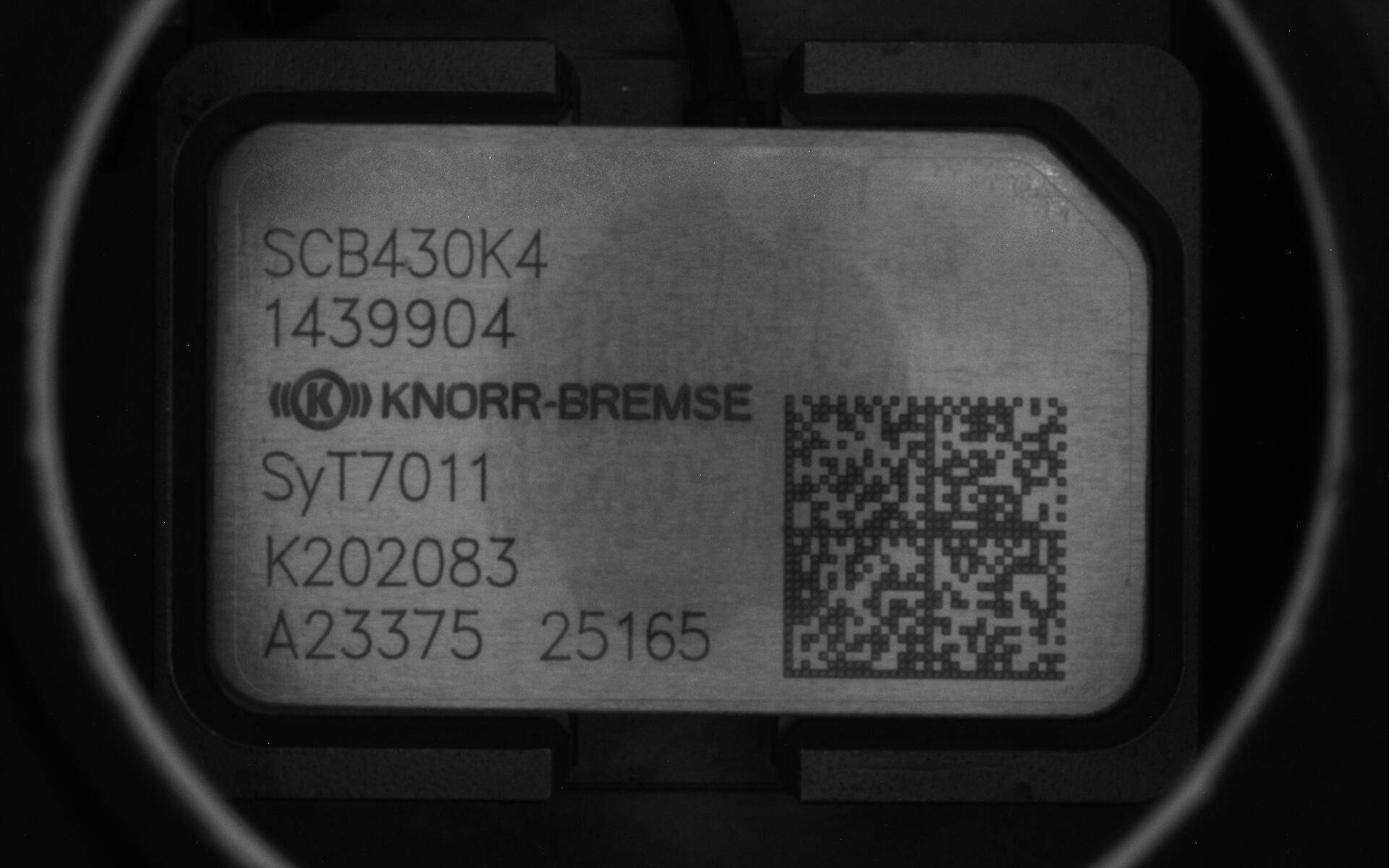}
        \caption{}
        \label{fig:darker_image}
    \end{subfigure}
    \hspace{0.5em} 
    \begin{subfigure}[t]{0.45\textwidth}
        \centering
        \includegraphics[width=\textwidth]{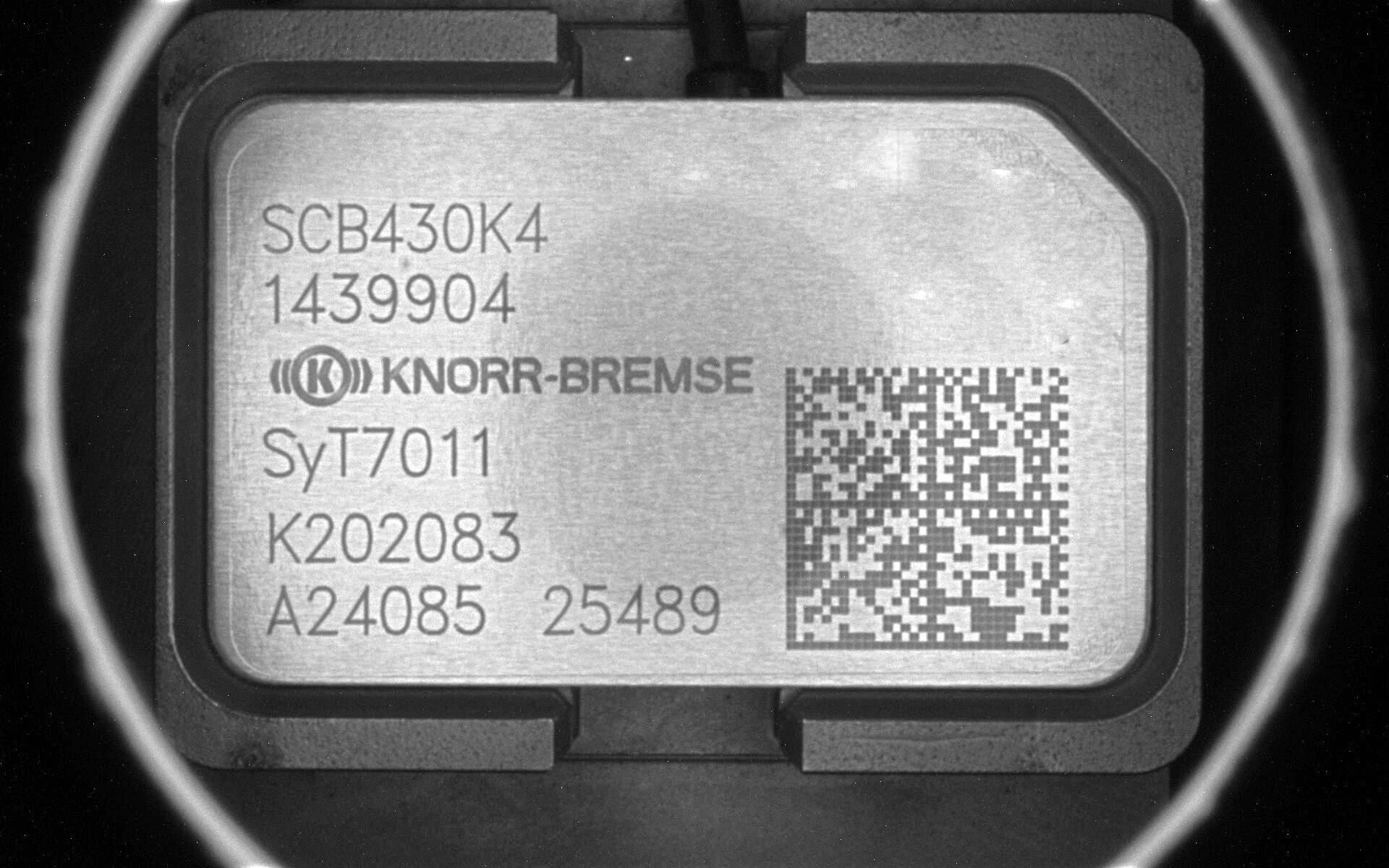}
        \caption{}
        \label{fig:brigher_image}
    \end{subfigure}
    \caption{
        Variation in lighting conditions: (a) darker image and (b) brighter image.
    }
    \label{fig:lighting_variation}
\end{figure}

The images in Figure~\ref{fig:lighting_variation} demonstrate variations in lighting conditions, which are commonly encountered in real-world production environments. Figure~\ref{fig:darker_image} represents a darker image captured under low-light conditions, while Figure~\ref{fig:brigher_image} represents a brighter image captured under high-light conditions. These variations in lighting are used to evaluate the robustness of the proposed pipeline and ensure its performance under diverse environmental conditions.

\begin{figure}[h!]
    \centering
    \begin{subfigure}[t]{0.45\textwidth}
        \centering
        \includegraphics[width=\textwidth]{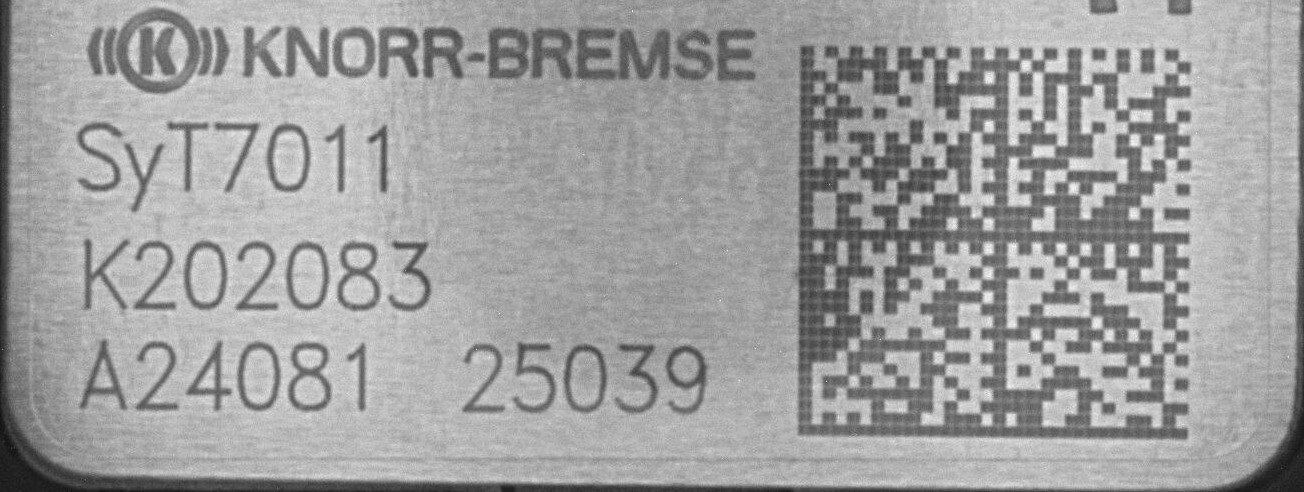}
        \caption{}
        \label{fig:variation1}
    \end{subfigure}
    \hspace{0.5em} 
    \begin{subfigure}[t]{0.45\textwidth}
        \centering
        \includegraphics[width=\textwidth]{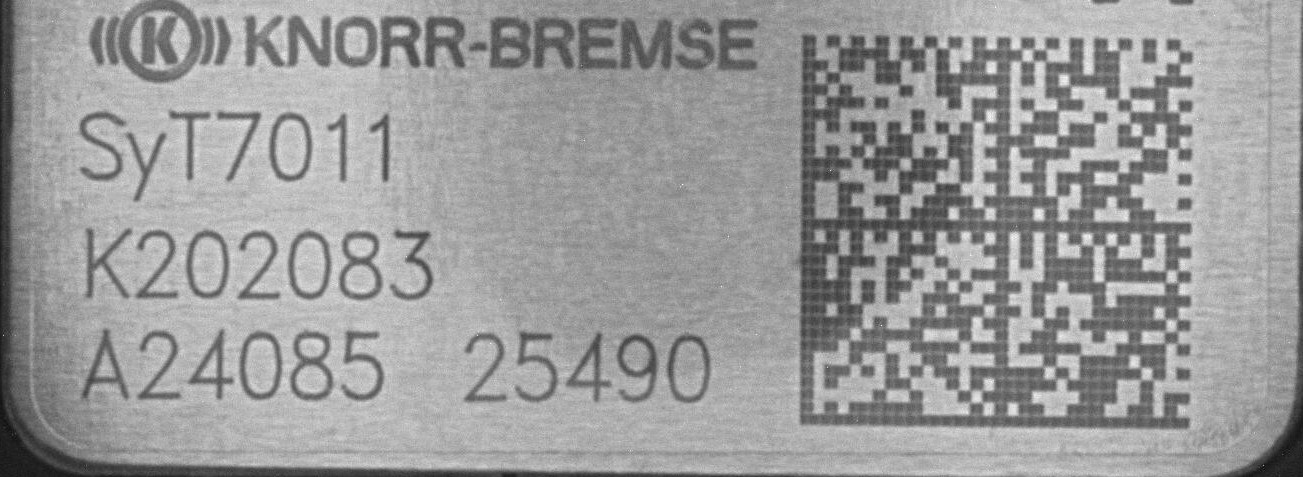}
        \caption{}
        \label{fig:variation2}
    \end{subfigure}
    \vspace{0.5em}
    \begin{subfigure}[t]{0.45\textwidth}
        \centering
        \includegraphics[width=\textwidth]{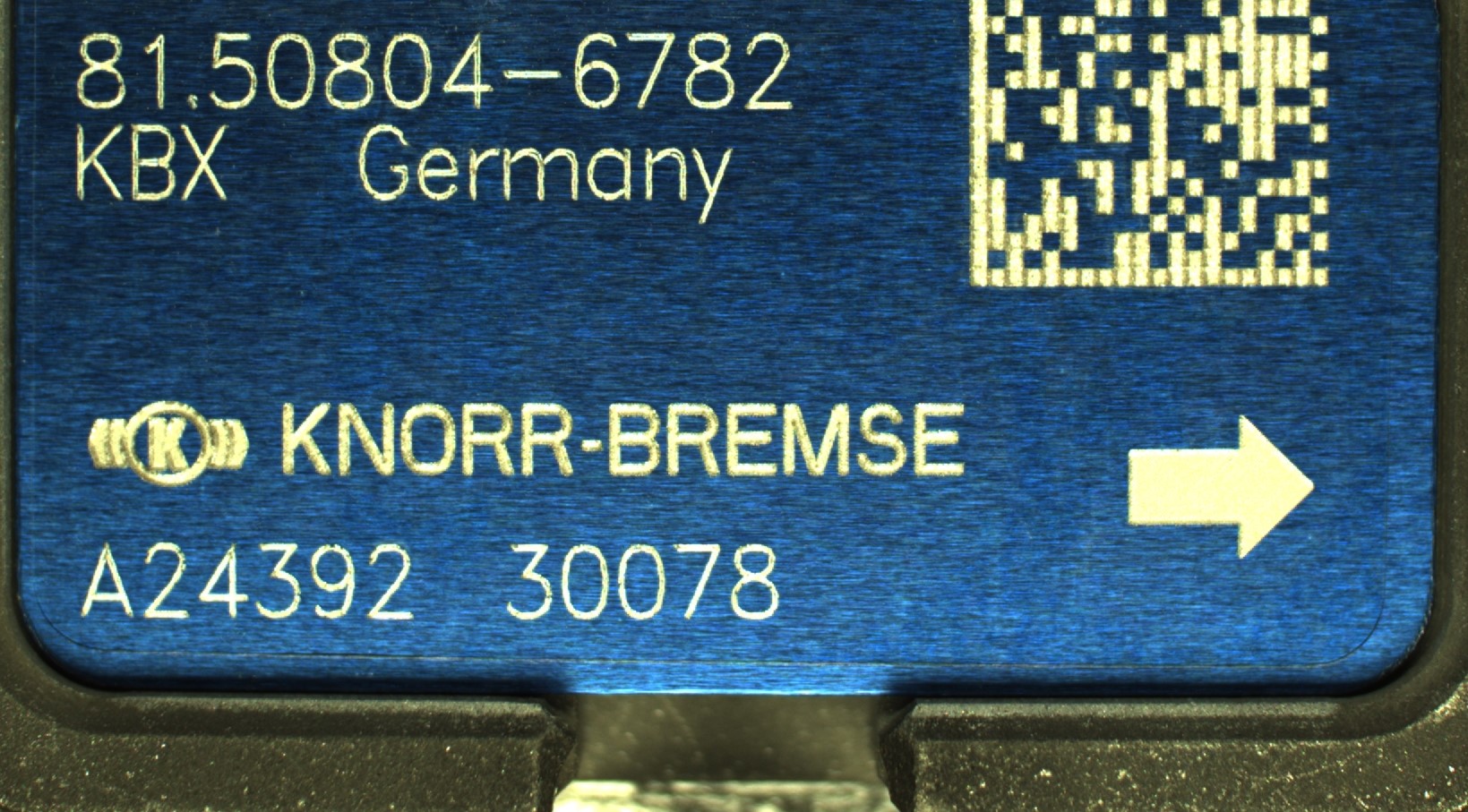}
        \caption{}
        \label{fig:color_variation}
    \end{subfigure}
    \caption{
        (a) and (b) demonstrate variations in the string content of nameplates
        (c) shows a different customer nameplate with variations in content and layout
    }
    \label{fig:heterogeneity_data}
\end{figure}

The images above depict the lower half of nameplates, with certain details omitted to maintain confidentiality. Figure~\ref{fig:variation1} and Figure~\ref{fig:variation2} illustrate variations in the string content of nameplates, showcasing the heterogeneity of the dataset. Figure~\ref{fig:color_variation} presents a nameplate from a different customer, where the content and layout differ significantly.

\begin{figure}[H]
    \centering
    \begin{subfigure}[t]{0.08\textwidth}
        \includegraphics[width=\linewidth]{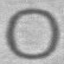}
    \end{subfigure} \hspace{0.2em}
    \begin{subfigure}[t]{0.08\textwidth}
        \includegraphics[width=\linewidth]{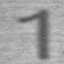}
    \end{subfigure} \hspace{0.2em}
    \begin{subfigure}[t]{0.08\textwidth}
        \includegraphics[width=\linewidth]{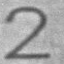}
    \end{subfigure} \hspace{0.2em}
    \begin{subfigure}[t]{0.08\textwidth}
        \includegraphics[width=\linewidth]{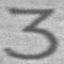}
    \end{subfigure} \hspace{0.2em}
    \begin{subfigure}[t]{0.08\textwidth}
        \includegraphics[width=\linewidth]{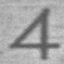}
    \end{subfigure} \hspace{0.2em}
    \begin{subfigure}[t]{0.08\textwidth}
        \includegraphics[width=\linewidth]{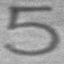}
    \end{subfigure}
    
    \vspace{0.5em} 

    \begin{subfigure}[t]{0.08\textwidth}
        \includegraphics[width=\linewidth]{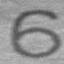}
    \end{subfigure} \hspace{0.2em}
    \begin{subfigure}[t]{0.08\textwidth}
        \includegraphics[width=\linewidth]{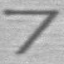}
    \end{subfigure} \hspace{0.2em}
    \begin{subfigure}[t]{0.08\textwidth}
        \includegraphics[width=\linewidth]{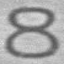}
    \end{subfigure} \hspace{0.2em}
    \begin{subfigure}[t]{0.08\textwidth}
        \includegraphics[width=\linewidth]{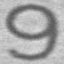}
    \end{subfigure} \hspace{0.2em}
    \begin{subfigure}[t]{0.08\textwidth}
        \includegraphics[width=\linewidth]{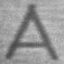}
    \end{subfigure} \hspace{0.2em}
    \begin{subfigure}[t]{0.08\textwidth}
        \includegraphics[width=\linewidth]{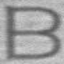}
    \end{subfigure}
    
    \vspace{0.5em} 

    \begin{subfigure}[t]{0.08\textwidth}
        \includegraphics[width=\linewidth]{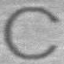}
    \end{subfigure} \hspace{0.2em}
    \begin{subfigure}[t]{0.08\textwidth}
        \includegraphics[width=\linewidth]{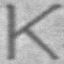}
    \end{subfigure} \hspace{0.2em}
    \begin{subfigure}[t]{0.08\textwidth}
        \includegraphics[width=\linewidth]{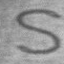}
    \end{subfigure} \hspace{0.2em}
    \begin{subfigure}[t]{0.08\textwidth}
        \includegraphics[width=\linewidth]{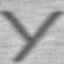}
    \end{subfigure} \hspace{0.2em}
    \begin{subfigure}[t]{0.08\textwidth}
        \includegraphics[width=\linewidth]{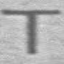}
    \end{subfigure} \hspace{0.2em}

    \caption{Example character images used for training and testing the ResVAE model}
    \label{fig:character_example}
\end{figure}

Figure~\ref{fig:character_example} shows the character images used to train and test the ResVAE model. The dataset includes numbers from 0 to 9 and specific letters (S, C, B, K, Y, T, A), selected because they appear in specific customer nameplate content. 

\begin{figure}[H]
    \centering
    \begin{subfigure}[t]{0.08\textwidth}
        \includegraphics[width=\linewidth]{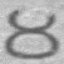}
    \end{subfigure} \hspace{0.2em}
    \begin{subfigure}[t]{0.08\textwidth}
        \includegraphics[width=\linewidth]{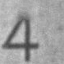}
    \end{subfigure} \hspace{0.2em}
    \begin{subfigure}[t]{0.08\textwidth}
        \includegraphics[width=\linewidth]{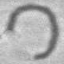}
    \end{subfigure} \hspace{0.2em}
    \begin{subfigure}[t]{0.08\textwidth}
        \includegraphics[width=\linewidth]{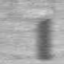}
    \end{subfigure} \hspace{0.2em}
    \begin{subfigure}[t]{0.08\textwidth}
        \includegraphics[width=\linewidth]{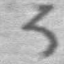}
    \end{subfigure} \hspace{0.2em}
    \begin{subfigure}[t]{0.08\textwidth}
        \includegraphics[width=\linewidth]{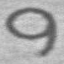}
    \end{subfigure}
    
    \vspace{0.5em} 

    \begin{subfigure}[t]{0.08\textwidth}
        \includegraphics[width=\linewidth]{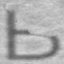}
    \end{subfigure} \hspace{0.2em}
    \begin{subfigure}[t]{0.08\textwidth}
        \includegraphics[width=\linewidth]{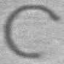}
    \end{subfigure} \hspace{0.2em}
    \begin{subfigure}[t]{0.08\textwidth}
        \includegraphics[width=\linewidth]{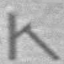}
    \end{subfigure} \hspace{0.2em}
    \begin{subfigure}[t]{0.08\textwidth}
        \includegraphics[width=\linewidth]{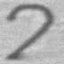}
    \end{subfigure} \hspace{0.2em}
    \begin{subfigure}[t]{0.08\textwidth}
        \includegraphics[width=\linewidth]{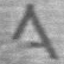}
    \end{subfigure} 
    \caption{Examples of synthetically generated defective characters}
    \label{fig:manu_character_example}
\end{figure}

Due to the limited availability of naturally occurring defective character images, synthetic defects were introduced to the dataset using the GIMP image manipulation tool \cite{cutts1997introduction}. Figure~\ref{fig:manu_character_example} illustrates examples of manually altered characters, such as distorted edges, incomplete strokes, or misaligned components. These manipulated images were used to enhance the diversity and increase the size of the training and testing data.

\begin{figure}[h!]
    \centering
    \begin{subfigure}[t]{0.45\textwidth}
        \centering
        \includegraphics[width=\textwidth]{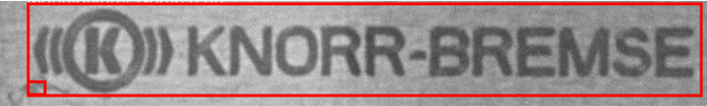}
    \end{subfigure}
    \caption{
        Misclassification of a surface scratch as a defect in the logo defect detection module 
    }
    \label{fig:logo_scracth}
\end{figure}

Figure~\ref{fig:logo_scracth} illustrates a surface scratch on the nameplate, which is not considered a defect based on the inspection criteria. However, the logo defect detection module incorrectly detects it as a defect and highlights the affected region in red.  
\end{appendices}

\end{document}